\documentclass[10pt,twocolumn,letterpaper]{article}

\usepackage{cvpr}      % To produce the REVIEW version
\DeclareMathOperator*{\argmin}{arg\,min}

\definecolor{cvprblue}{rgb}{0.21,0.49,0.74}
\usepackage{booktabs}
\usepackage{marvosym}
\usepackage{multirow}
\usepackage{subcaption}
\usepackage[pagebackref,breaklinks,colorlinks,allcolors=cvprblue]{hyperref}

\newcommand{\modelname}{{BrainCoDec}}

\def\paperID{9169} % *** Enter the Paper ID here
\def\confName{CVPR}
\def\confYear{2026}

\title{Meta-learning In-Context Enables Training-Free Cross Subject Brain Decoding}

\author{
Mu Nan$^{1, 2 *}$
\and 
Muquan Yu$^{1, 3 *}$
\and 
Weijian Mai$^{1, 4}$
\and 
Jacob S. Prince$^{5}$
\and 
Hossein Adeli$^{6}$
\and 
Rui Zhang$^{1}$
\and 
Jiahang Cao$^{1, 4}$
\and 
Benjamin Becker$^{1}$
\and 
John A. Pyles$^{7}$
\and 
Margaret M. Henderson$^{8}$
\and 
Chunfeng Song$^{4}$
\and
Nikolaus Kriegeskorte$^{6}$
\and 
Michael J. Tarr$^{8}$
\and 
Xiaoqing Hu$^{1}$
\and 
Andrew F. Luo$^{1 \mbox{\scriptsize \Letter}}$
\and
\\
$^1$University of Hong Kong \hspace{1.5em} 
$^2$Shenzhen Loop Area Institute \hspace{1.5em}
$^3$Chinese University of Hong Kong \\
$^4$Shanghai Artificial Intelligence Laboratory \hspace{1.5em} 
$^5$Harvard University \hspace{1.5em} 
$^6$Columbia University \\ 
$^7$University of Washington \hspace{1.5em} 
$^8$Carnegie Mellon University \\
{\tt\small ezacngmpg@connect.hku.hk, mqyu@link.cuhk.edu.hk, aluo@hku.hk}
}

\begin{document}
\maketitle
\begin{abstract}
Visual decoding from brain signals is a key challenge at the intersection of computer vision and neuroscience, requiring methods that bridge neural representations and computational models of vision. A field-wide goal is to achieve generalizable, cross-subject models. A major obstacle towards this goal is the substantial variability in neural representations across individuals, which has so far required training bespoke models or fine-tuning separately for each subject. To address this challenge, we introduce a meta-optimized approach for semantic visual decoding from fMRI that \textbf{generalizes to novel subjects without any fine-tuning}. By simply conditioning on a small set of image-brain activation examples from the new individual, our model rapidly infers their unique neural encoding patterns to facilitate robust and efficient visual decoding. Our approach is explicitly optimized for in-context learning of the new subject's encoding model and performs decoding by hierarchical inference, inverting the encoder. First, for multiple brain regions, we estimate the per-voxel visual response encoder parameters by constructing a context over multiple stimuli and responses. Second, we construct a context consisting of encoder parameters and response values over multiple voxels to perform aggregated functional inversion. We demonstrate strong cross-subject and cross-scanner generalization across diverse visual backbones without retraining or fine-tuning. Moreover, our approach requires neither anatomical alignment nor stimulus overlap. This work is a critical step towards a generalizable foundation model for non-invasive brain decoding. Code and models are publicly available at \href{https://github.com/ezacngm/brainCodec}{https://github.com/ezacngm/brainCodec}. 
% \footnotes{$*$: Equal contribution, \Letter: Corresponding author.}
% All code is available~\href{https://anonymous.4open.science/r/braInCodec-7079/}{here}.
\end{abstract}

\section{Introduction}
\label{sec:intro}
\begin{figure*}[t]
    \centering
    \vspace{-1em}\includegraphics[width=0.8\linewidth]{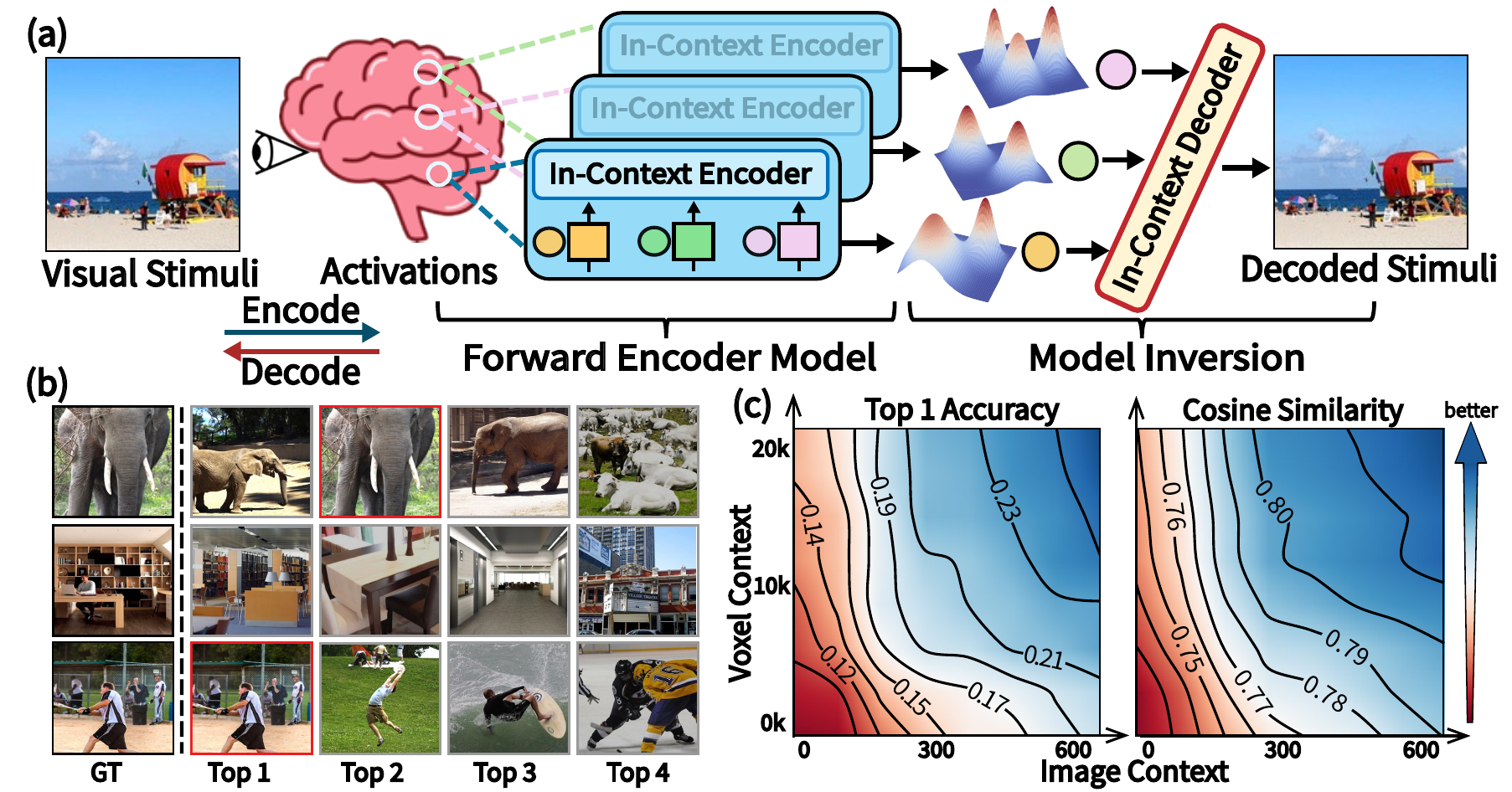}
    \caption{\textbf{Overview of our hierarchical brain decoding framework.} Encoders predict brain activity from stimulus, while decoders reconstruct stimulus from brain activity. \textbf{(a)} Our framework can generalize to novel subjects \textbf{without any fine-tuning}. In the first stage, we infer parameters of a forward model (image-computable encoder)  by constructing a context using stimuli/activity pairs for a single voxel, repeated for every voxel. In the second stage, we construct a context across multiple voxels, fusing the encoder parameters with observed brain activations to decode the stimuli. Our approach requires neither anatomical alignment, nor stimulus overlap. \textbf{(b)} Decoding results on BOLD5000 after training on NSD, our method can \textbf{generalize without fine-tuning across scanners, voxel size, and subjects}. \textbf{(c)} Our model performance positively scales with both the number of images provided in Stage 1, and the number of voxels provided in Stage 2.}
    \label{fig:1_teaser}
    \vspace{-10pt}
\end{figure*}
Developing robust theories of intelligence requires generalizable, population-wide models of human brain function. An important step has been the development of high-fidelity visual decoders of brain activity~\citep{takagi2023high,chen2023seeing}, enabled by conditional image generation models and the availability of high-quality fMRI visual datasets. Visual reconstruction serves as a unique and demanding testbed for conditional generation, requiring vision models to synthesize images from signals that are not only noisy but also highly abstract. A common strategy decomposes this challenge into two sub-problems: \textbf{(1)} learning a mapping from high-dimensional brain activity to a compact visual-semantic representation; and \textbf{(2)} synthesizing naturalistic images from that representation. The synthesis challenge has been addressed by leveraging large-scale generative models as image priors~\cite{rombach2022high,ramesh2022hierarchical}. Simultaneously, high-quality neural activity datasets~\citep{horikawa2017generic,chang2019bold5000,allen2022massive,gong2023large,hebart2023things,li2025triple} at scale have provided sufficient data to solve the mapping sub-problem on an individual basis, driving the recent surge in high-fidelity, within-subject reconstructions.

Despite this recent progress, a critical barrier prevents widespread application of brain decoding: current models \textbf{cannot generalize across subjects}, necessitating per-subject models or subject-specific fine-tuning~\cite{scotti2024mindeye2,wang2024mindbridge}. This challenge is rooted in the profound inter-subject variability in neural signals which arises from complex interacting sources~\cite{van2008individual}, including differences in anatomical structure and functional organization shaped by development, individual experience, and neuroplasticity~\citep{tarr2000ffa,gauthier2000expertise,willems2010cerebral,cai2013complementary}. As a result, the mapping function learned for one individual is ineffective for another, necessitating retraining or fine-tuning via gradient descent, a data-intensive and computationally demanding process. 
% This subject specificity limits the extent to which models can generalize across the population and constrains their scalability and practical deployment. 
Developing a data-efficient generalizable cross-subject visual decoding model is therefore essential for building population-wide theories and for enabling applications in brain-computer interfaces (BCIs), cognitive assessment, and personalized diagnostics.

A principled approach is to recognize that neural decoding is fundamentally an inverse problem. A robust solution should be constrained by an accurate forward model of the system that characterizes how the brain of an individual subject represents information. In computational neuroscience, this forward model is referred to as an ``encoding model''~\cite{naselaris2011encoding}, which predicts brain activity from stimuli. Meanwhile, the inverse operation is performed by the decoding model. Following this principle, our approach structures the decoding process as a functional inversion problem that we solve hierarchically. \textbf{First}, we estimate the visual response function weights for individual voxels in-context~\cite{yu2025meta}; \textbf{Second}, we build a decoder that performs contextual integration across multiple brain regions to perform a subject-specific functional inversion to reconstruct the visual stimulus. This two-stage in-context learning process enables generalization to novel subjects \textbf{\underline{without any fine-tuning}} and \textbf{\underline{with relatively small amounts of new data}}. Since image synthesis from brain activity has been well explored using pretrained generative models, we instead focus on decoding image embeddings from novel subjects. 

We name our method BrainCoDec (Brain In-Context Decoding), and outline the  approach in Figure~\ref{fig:1_teaser}. Concretely: \textbf{(1)} Our method generalizes to novel subjects, requires no anatomical alignment or stimulus overlap, and is the first to work across different scanners and acquisition protocols without gradient-based finetuning. \textbf{(2)} Through selective dropout of functionally specialized regions and by using only a small subset of voxels from higher visual cortex, we demonstrate strong robustness to input variability. \textbf{(3)} Attention visualizations across images from diverse categories reveal interpretable spatial maps that align closely with known functional regions of the visual cortex. This approach marks a significant step towards a truly universal and scalable brain foundation model for investigating neural representations across the human population.

\section{Related work}
\vspace{-.2cm}
\begin{figure*}[th]
    \centering
    \includegraphics[width=0.8\linewidth]{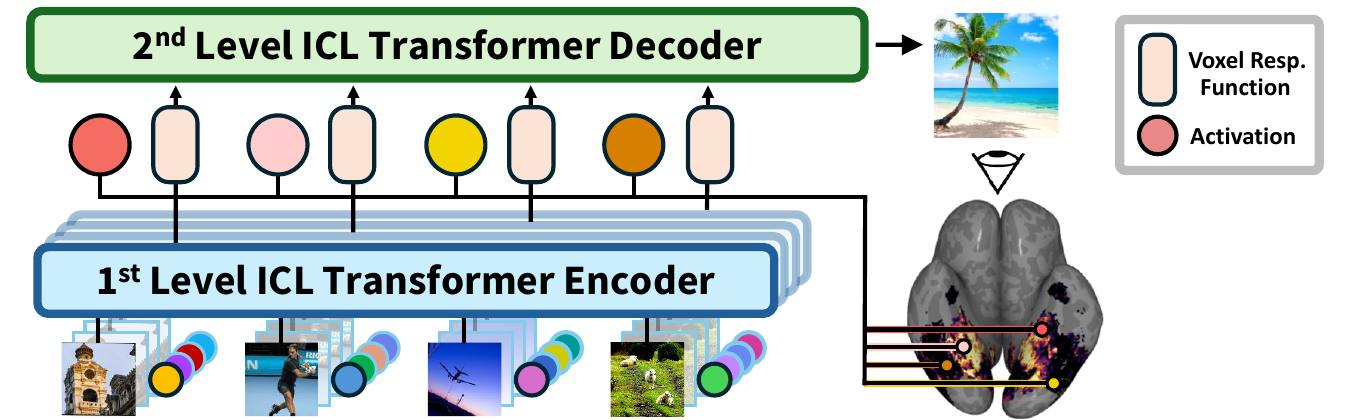}
    \caption{\textbf{Model architecture of \modelname.} In stage one, the in-context encoder infers encoder parameters by in-context learning across stimuli/activation pairs for a single voxel. This is repeated across the voxels of interest. In stage two, we integrate across multiple voxels, taking as input the voxelwise parameters and activation corresponding to a novel image. Both stages can vary the context sizes.}
    \vspace{-15pt}
    \label{fig:2_model_archi}
\end{figure*}
\noindent\textbf{Computational Encoding and Decoding Models.} 
Computational analysis of neural data usually leverage two complementary approaches. Encoding models predict neural activity from stimuli, and decoding models that reconstruct stimuli from brain activity~\cite{naselaris2011encoding,kamitani2005decoding,norman2006beyond, han2019variational,seeliger2018generative,shen2019deep,ren2021reconstructing,dai2025mindaligner,gifford2024opportunities,mai2024brain}. Both approaches have benefited from the development from feature extractors trained on large-scale datasets, with the dominant approach leveraging linear mappings from learned features to neural activity~\citep{dumoulin2008population,gucclu2015deep,klindt2017neural,eickenberg2017seeing,wen2018neural,gaziv2022self}, with more recent approaches utilizing attention based parameterization~\citep{adeli2023predicting,beliy2024wisdom,bao2025mindsimulator}. Core to our current work is the approach proposed in~\cite{yu2025meta}, which meta-optimizes an encoding model to generalize to novel subjects. Encoders can be used to investigate the selectivity in visual cortex~\citep{khosla2022high,khosla2022characterizing,efird2024s,yang2024brain,yang2024alignedcut,luo2024brain,Sarch2023.05.29.542635,lappe2024parallel}, or combined with generative models to synthesize new stimuli~\citep{walker2019inception,bashivan2019neural,ponce2019evolving,ratan2021computational,gu2022neurogen,pierzchlewicz2023energy,luo2023brain,cerdas2024brainactiv,luo2024brainscuba,matsuyama2025lavca}. By leveraging generative models, stimulus can be decoded from fMRI, EEG, and MEG for images~\citep{takagi2023high,chen2023seeing,lu2023minddiffuser,ozcelik2023brain,doerig2022semantic,ferrante2023brain,liu2023brainclip,mai2023unibrain,scotti2024mindeye2,benchetrit2023brain,li2024visual,guo2025neuro,mai2025synbrain}, dynamic visual stimuli~\citep{zhu2025multi,schneider2023learnable,chen2023cinematic,gong2024neuroclips,yeung2024neural,liu2024eeg2video,fosco2024brain}, and speech/audio/language~\citep{pasley2012reconstructing,varoquaux2017assessing,bellier2023music,oota2023speech,jo2024eeg,willett2023high,metzger2023high}. Recent work seeks to achieve generalization via flatmaps~\citep{lane2025scaling,wang2025zebra}, $1D$ pooling~\citep{wang2024mindbridge} or surface learning ~\citep{cui2025brainx}, these approaches implicitly (flatmaps \& pooling) or explicitly (surface) require anatomical alignment. 
% \vspace{-.2cm}

\noindent\textbf{Inverting Encoding Models for Decoding.} Prior work has sought to decode (identify the category or semantic nearest neighbor) of stimuli by comparing patterns of neural activations~\cite{haxby2001distributed,o2000mental,haynes2006decoding,kok2014shape}. Reconstructing viewed images from neural activity by inverting a forward model (encoder) has been previously demonstrated using simple stimuli~\cite{brouwer2009decoding}, which inverts the encoder using ordinary least squares to solve for the color of the image. Similar approaches that convert between encoders that predict neural activation and decoders have been utilized in the context of motion direction~\cite{kok2013prior,saproo2014attention},  orientation~\cite{brouwer2011cross}, and more complex stimuli like faces~\cite{haufe2014interpretation, cowen2014neural}, natural images~\cite{kay2008identifying, naselaris2009bayesian, schoenmakers2013linear} and movies~\cite{nishimoto2011reconstructing}. Generally these methods are based on the principle of matching stimuli and their predicted brain activity to the true observed brain activity, and decoding the stimuli by solving or identifying the solution. Our learned approach significantly extends this prior work by functioning even when the system is under-determined (fewer voxels than stimulus representation), and being able to account for biases in the encoder estimation. 

\noindent\textbf{Meta-Learning and In-Context Learning.} Meta-learning focuses on training models to rapidly adapt to new tasks by leveraging prior knowledge acquired from a distribution of related tasks~\cite{hospedales2021meta}. It facilitates fast generalization to novel problems with few examples and minimal training effort. Classic approaches include meta‑optimization methods~\citep{finn2017model,nichol2018first,rajeswaran2019meta} and metric-based formulations~\citep{snell2017prototypical}. In parallel, large language models display strong in‑context learning (ICL) capability~\citep{brown2020language,von2023transformers}: given prompts with demonstrations, model behaviors could be adjusted at inference time effectively without updating parameters~\citep{min2021metaicl,coda2023meta}. These observations may suggest that in-context learning serves as an implicit meta-learning mechanism, whereby transformers develop internal adaptation procedures during the pretraining stage~\citep{garg2022can,dai2022can}. In our work, which aims to learn the functional mapping between visual stimuli and voxelwise brain responses, we construct a framework that integrates meta-training with in-context learning. This approach enables training-free adaptation to novel subjects.
\begin{figure*}[ht]
    \centering
    \includegraphics[width=0.75\linewidth]{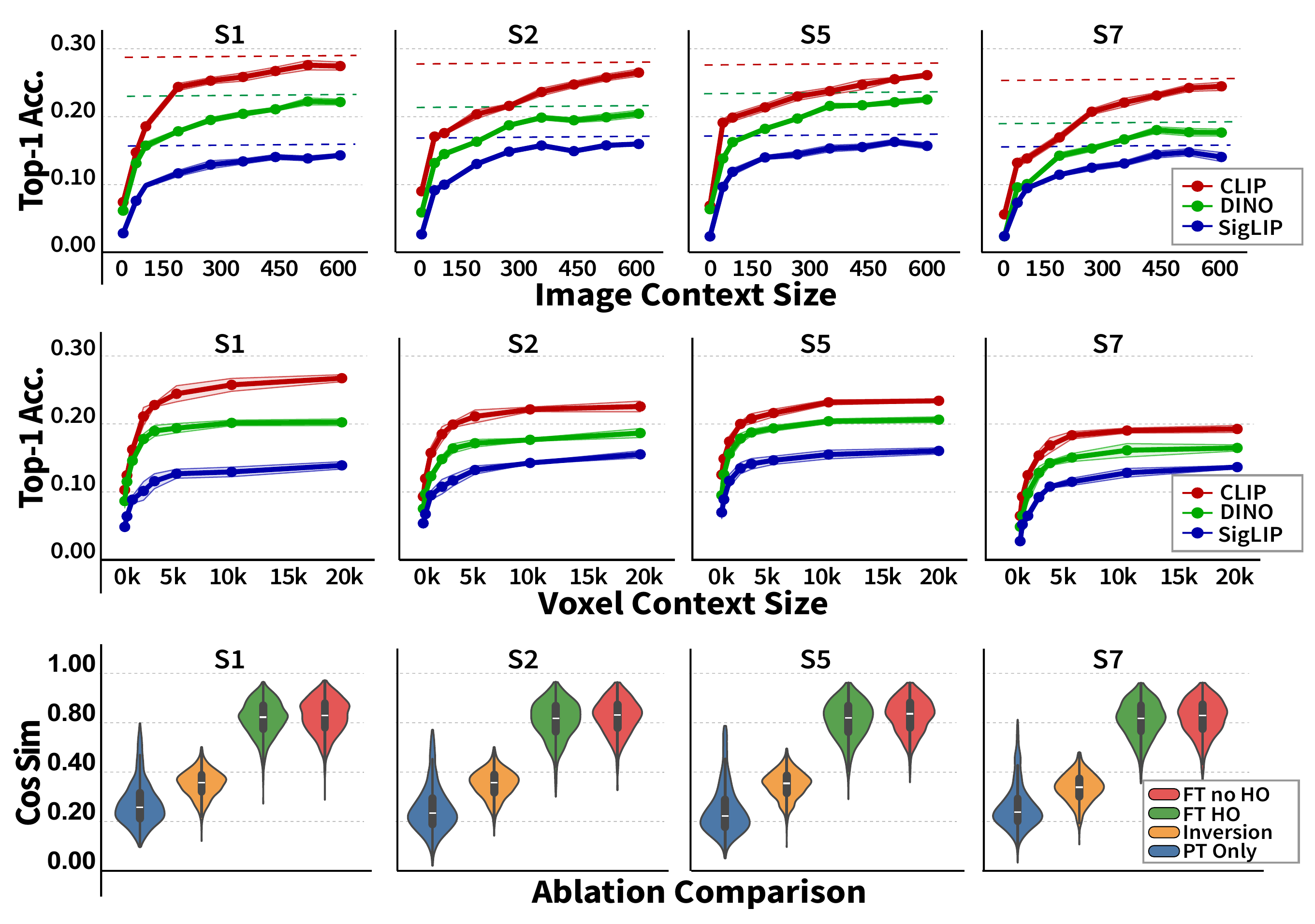}
    \caption{
    \textbf{Contextual scaling and ablation analysis of \modelname.} 
    \textbf{Top:} Image-context scaling from stage 1. Decoder performance scales positively with more images collected for the novel subject. \textbf{Middle:} Voxel-context scaling from stage 2. Top-1 retrieval accuracy improves consistently as the number of in-context voxels increases with all visual backbones across all subjects. 
    \textbf{Bottom:} Ablation comparison. Cosine similarities for four variants using CLIP backbone, synthetic data pretraining  (PT Only), gradient inversion (Inversion), training with subject hold-out (FT~HO; \modelname), and training on seen subjects (FT~no~HO). These results show that models trained with real neural data outperform the models trained with only synthetic data, with only marginal gains from fine-tuning on a subject. }
    \label{fig:3_context_scaling}
    \vspace{-15pt}
\end{figure*}
\section{Methods}
Our method is based on the learned inversion of a set of encoders. The framework leverages meta-learning, and uses few-shot, in-context examples for the decoding of unseen stimuli (Figure~\ref{fig:2_model_archi}). For unseen subjects, this approach does not require \underline{\textbf{any fine-tuning}}. We first define the problem in Section~\ref{motivation}, and discuss how stimuli can be recovered by inverting a set of encoders in Section~\ref{inversion}. In Section~\ref{hierarchical} we discuss how  hierarchical in-context learning can enable training-free decoding on novel subjects. Since image generation is relatively well studied, in this work we focus on decoding an image embedding, as it is core to the mapping problem, and evaluate method performance using retrieval following~\cite{kong2024toward,xia2024umbrae,scotti2024mindeye2}.

\subsection{Motivation and Problem Definition}
\label{motivation}
Substantial cross-subject variability in neural responses poses a major obstacle to generalizable brain decoding. Rather than directly learning a fixed inversion mapping, we reformulate neural decoding as a meta-learning problem that learns \underline{\textbf{how to perform functional inversion}}. Crucially, our approach does not rely on any shared stimuli or anatomical alignment across subjects. 

Formally, let an image $I$ be represented by its embedding vector $\mathcal{I} = \phi(I) \in \mathbb{R}^{1\times d}$, where $\phi$ denotes a pretrained image feature extractor such as CLIP~\cite{radford2021learning}, and $d$ is the embedding dimension. For a given image stimulus $I$, the corresponding fMRI response for a subject is denoted as $B_{\mathcal{I}} = (\beta_1, \beta_2, \dots, \beta_{K})_{\mathcal{I}} \in \mathbb{R}^{1\times K}$, where $K$ is the number of voxels in the subject’s visual cortex. During testing, for a new subject we observe a small set of $n$ context image-brain activation pairs $\{(\mathcal{I}_i, B'_i)\}_{i=1}^n$, where $B'_i$ represents the measured voxel activations for the $i$-th image. Our goal is to infer the embedding $\mathcal{I}_{\text{novel}}$ of an unseen image from its corresponding brain response $B'_{\text{novel}}$ using only these context examples.

\subsection{Decoding as the Functional Inversion}
\label{inversion}
Let us assume that the forward model (image-computable encoder) predicts for a given voxel $v_k$: $f_k(\mathcal{I})\Rightarrow \hat{\beta}_{\mathcal{I},k}$. Ideally, given a sufficient number of voxels $\{v_1, v_2, v_3, \dots, v_j\}$ where $j \gg d$ and encoder functions that are error free, we can uniquely solve for the stimulus $\mathcal{I}^*$ by inverting the encoding model such that: 
\vspace{-0.3cm}
\begin{align}
    \mathcal{I}^* = \argmin_\mathcal{I} \left(\sum_{m}^{j}\|f_m(\mathcal{I})-\beta_{m}\|_2^2\right)
\end{align}
In practice, the forward models of the encoders could be biased and inaccurate, the choice of metric/distance may affect the solution, and knowledge about the distribution of the inputs or outputs may improve the decoder. Unlike prior work that learn decoders to map from neural representations to stimuli directly, our approach takes a meta-learning view and learns a model to perform in-context functional inversion across a variable number of higher visual cortex voxels.

%  Inspired by BraInCoRL~\cite{yu2025meta}, which meta-learns voxel-wise response functions $f_{\omega,k}$ from a few contextual examples, we extend this idea hierarchically. In the first stage, a pretrained BraInCoRL model estimates the voxel-specific encoding functions $f_{\omega,k}$. In the second stage, our decoder performs an \textit{aggregated functional inversion} defined as:
% \begin{equation}
% F^{-1}(B'_{\mathcal{I}}) \approx \mathcal{A}\left(\{f_{\omega,k}^{-1}(\beta_{\mathcal{I},k})\}_{k=1}^K\right)
% \end{equation}
% where $\mathcal{A}(\cdot)$ denotes a learned aggregation operator that integrates voxel-level inversions into a coherent subject-specific prediction. This hierarchical design enables robust cross-subject generalization without gradient-based fine-tuning.
\begin{figure*}[ht]
    \centering
    \includegraphics[width=0.75\linewidth]{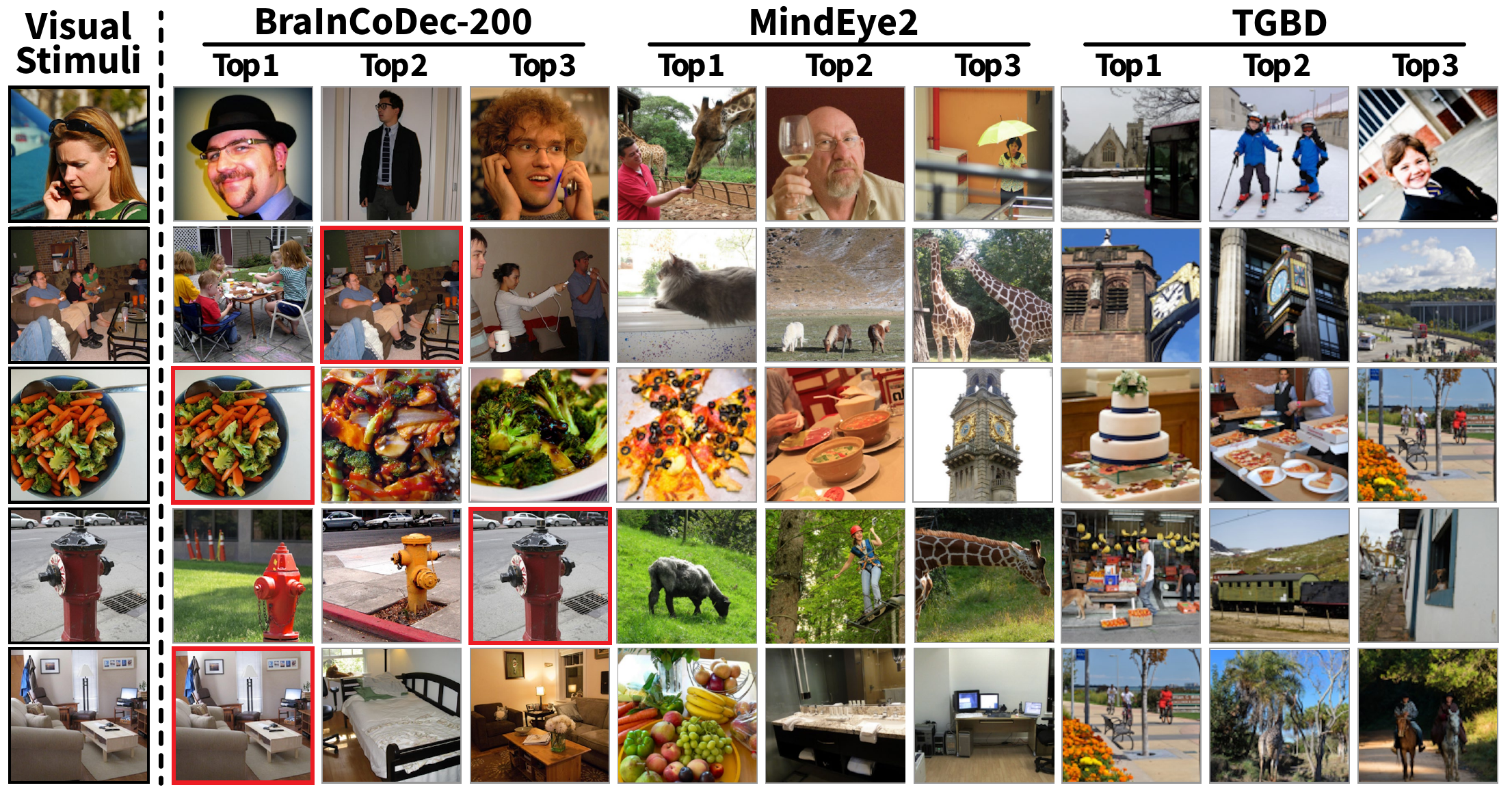}
    \caption{
    \textbf{Image retrieval comparison on a subject unseen during training (S1)}. For each method (\modelname-200, MindEye2 + anatomical alignment, TGBD), columns list the Top-$1$–$3$ retrieved images out of $907$ test images from left to right, ranked by similarity in the evaluation embedding space. Red boxes mark correct hits. Our model can yield very high semantic retrieval consistency without any fine-tuning.}
    \label{fig:4_retrieval_visualization}
    \vspace{-10pt}
\end{figure*}

\begin{table*}[t]
\centering
\caption{
\textbf{Quantitative comparison of \underline{unseen} subject brain decoding performance.} 
Top-1 and Top-5 retrieval accuracy (\%) on unseen NSD subjects (S1, S2, S5, S7) for MindEye2 + anatomical alignment, TGBD, and \modelname-200 (200 in-context images). Our method substantially outperforms prior methods while requiring neither subject-specific fine-tuning nor large-scale training data. 
Mean accuracies across subjects are reported in the rightmost column; additional metrics and standard deviations are provided in the \textbf{Appendix}.
}
\renewcommand{\arraystretch}{0.9}
\setlength{\tabcolsep}{4pt}
\begin{tabular}{ccccccccccc}
\toprule
 & \multicolumn{2}{c}{\textbf{S1}} & \multicolumn{2}{c}{\textbf{S2}} & \multicolumn{2}{c}{\textbf{S5}} & \multicolumn{2}{c}{\textbf{S7}} & \multicolumn{2}{c}{\textbf{Mean}} \\ 
\cmidrule(lr){2-3} \cmidrule(lr){4-5} \cmidrule(lr){6-7} \cmidrule(lr){8-9} \cmidrule(lr){10-11}
\textbf{Models} & Top-1$\uparrow$ & Top-5$\uparrow$ & Top-1$\uparrow$ & Top-5$\uparrow$ & Top-1$\uparrow$ & Top-5$\uparrow$ & Top-1$\uparrow$ & Top-5$\uparrow$ & Top-1$\uparrow$ & Top-5$\uparrow$ \\
\midrule
MindEye2 \cite{scotti2024mindeye2} & 4.11\% & 12.9\% & 3.82\% & 10.70\% & 2.87\% & 9.58\% & 2.51\% & 6.49\% & 3.90\% & 9.81\% \\
TGBD \cite{kong2024toward}         & 1.27\% & 3.89\% & 0.56\% & 2.33\% & 0.84\% & 3.34\% & 0.39\% & 1.41\% & 0.82\% & 3.09\% \\
\midrule
\textbf{\modelname-200} & \textbf{25.5\%} & \textbf{56.6\%} & \textbf{22.9\%} & \textbf{52.4\%} & \textbf{23.2\%} & \textbf{55.8\%} & \textbf{19.2\%} & \textbf{51.2\%} & \textbf{22.7\%} & \textbf{54.0\%} \\
\bottomrule
\end{tabular}
\label{tab:1_nsd_comparison}
\vspace{-10pt}
\end{table*}
\begin{figure*}[ht]
    \centering
    \vspace{-0.5em}\includegraphics[width=0.75\linewidth]{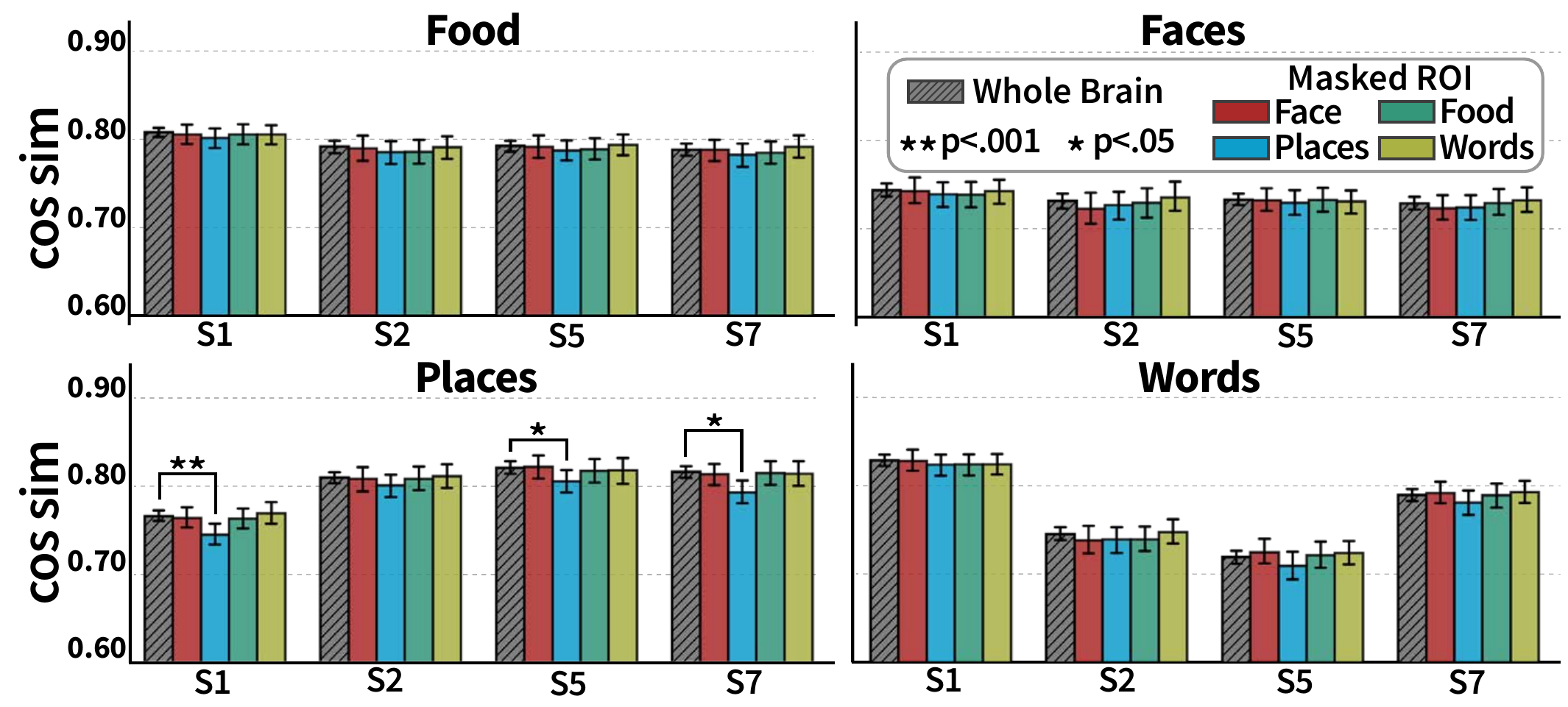}
    \vspace{-0.5em}
    \caption{    
    \textbf{Robustness of removing voxels from ROIs.}
    Cosine similarity of masking out category-specific voxels (Food, Faces, Places, Words) across four unseen NSD subjects on top-activating images from the test set. For each category, we compare performance using full context voxels from higher visual cortex versus masking out category-selective ROIs. Across nearly all conditions, masking the corresponding functional region has minimal impact on decoding performance, indicating strong robustness and distributed representation learning in \modelname. Masking scene-selective regions (PPA/OPA/RSC) leads to some performance drop.
    }
    \vspace{-15pt}
    \label{fig:5_roi_exp_maskedout}
\end{figure*}
\subsection{Hierarchical Training-Free Stimulus Decoding}
\label{hierarchical}
Our decoding approach leverages a hierarchical inference process, with two successive in-context stages, each with a distinct type of context. In \textbf{Stage 1}, we perform in-context inference across multiple stimulus-response pairs to infer the \underline{voxelwise response (encoder) function parameters}. We run this per-voxel, across all voxels of interest. In ~\textbf{Stage 2}, we construct a voxel context across multiple voxels to perform inversion and estimate the image embedding. Here the context consists of an aggregate of voxelwise encoder parameters and activations for a single novel stimulus. 

\noindent\textbf{Encoder Parameter Estimation.} 
In Stage~1 we adopt BrainCoRL's approach~\cite{yu2025meta} to estimate the per-voxel parameters. For a novel subject, for voxel $v_q$ we have a context defined by $\{(\mathcal{I}_1, \beta_{1,q}), (\mathcal{I}_2, \beta_{2,q}), ..., (\mathcal{I}_n, \beta_{n,q})\}$, where we have the voxel's activation in response to $n$ images. Let the pretrained BrainCoRL model be $T_\theta$, then:
\begin{equation}
\omega_q = T_\theta\left(\{(\mathcal{I}_t, \beta_{t,q})\}_{t=1}^{n}\right)
\end{equation}
where the model can output the voxelwise function weights of a novel subject without any fine-tuning. Note that we perform this stage independently for each voxel in higher visual cortex, computing contextual structure \textbf{\underline{across stimuli}} separately for each voxel.
%Note that we run this stage across every voxel in higher visual cortex independently, and the context here is constructed \textbf{\underline{across stimuli}} independently for each voxel. 

% This stage can estimate the encoder parameters for a novel subject without any fine-tuning.

% for a single subject we have a context set of $n$ paired observations $\{(\mathcal{I}_t, B'_t)\}_{t=1}^{n}$, where $\mathcal{I}_t$ denotes the image embedding and $B'_t = \{\beta_{k,t}\}_{k=1}^{m}$ ($m \leq K$) represents the corresponding fMRI responses recorded from a subset of voxels $\{v_k\}_{k=1}^{m}$ in the higher visual cortex. This stage constructs the \textbf{image context} over multiple image–brain pairs to infer voxel-specific response functions.
% For each voxel $v_k$, a pretrained voxel-wise response predictor $M$ infers its response parameter $\omega_k$ from the image context:

\noindent\textbf{Contextual Functional Inversion.} 
In Stage~2, the model performs functional inversion by constructing a context across voxels within a single subject. This approach allows us to flexibly adapt our model to novel subjects which have different voxel counts. Our approach does not require any reference to anatomy, and does not require cross-subject anatomical alignment. Each voxel $v_k$ is represented by a context token $c_k$, defined as the concatenation of its predicted response parameter $\omega_k$ derived from stage 1, 
and the measured activation $\beta_k$ from the novel stimulus, $c_k = [\omega_k, {\beta}_k]$. The voxel context for a subject is then $\{c_k\}_{k=1}^{m}$, where $m \leq K$. We train a transformer $P_\gamma$ with variable-length voxel contexts to approximate the aggregated inverse mapping:
{
\setlength{\abovedisplayskip}{5pt}
\setlength{\belowdisplayskip}{5pt}
\begin{equation}
\hat{\mathcal{I}} \approx P_\gamma(\{c_k\}_{k=1}^{m})
\end{equation}
}
where $P_\gamma$ denotes a learned transformer that jointly inverts the functional representations of multiple voxels.

\noindent\textbf{Test-time Context Scaling.}
At test time, when a new subject is presented, the number of $K$ voxels available for decoding may vary across individuals. This variability in context size poses a challenge for model generalization. Unlike transformers in language modeling, where outputs depend on the sequential order of tokens, our model should be invariant to both the number and the order of voxel token inputs. To accommodate variable-length contexts, we adopt logit scaling~\citep{su2021analyzing, chiang2022overcoming, bai2023qwen}. Assuming a query/key ($q,k$) with $d$ features and a length $l$ context:
{
\begin{align}
    \alpha_\text{orig} = \frac{q\cdot k}{\sqrt{d}}; \quad \alpha_\text{scaled} = \frac{\log{(l)} \cdot q \cdot k}{\sqrt{d}}
\end{align}
}

\noindent Our model integrates a \texttt{[CLS]} token for output. We omit positional embeddings to achieve order invariance.

\noindent\textbf{Training Objective.} 
To achieve both fine-grained alignment and instance-level discriminability, we employ a hybrid cosine-contrastive loss that combines cosine embedding loss and an InfoNCE loss. Let $\mathcal{I}$ be unit vectors:
{
\setlength{\abovedisplayskip}{4pt}
\setlength{\belowdisplayskip}{4pt}
\small
\begin{align}
    \mathcal{L}_\text{total} = \Big( \mathcal{L}_{\text{cos}} + \alpha \, \mathcal{L}_{\text{infoNCE}}\Big)
\end{align}
} 
where for a batch size of $N$:

\resizebox{.95\linewidth}{!}{
  \begin{minipage}{\linewidth}
  \begin{align*}
  \mathcal{L}_\text{total} = \frac{1}{N} \sum_{i=1}^N \times \left(\left(1 - \hat{\mathcal I}_i^\top \mathcal I_i\right) -
       \log \frac{\exp(\hat{\mathcal I}_i^\top \mathcal I_i/\tau)}
                     {\sum_{j=1}^N \exp(\hat{\mathcal I}_i^\top \mathcal I_j/\tau)}\right)
  \end{align*}
  \end{minipage}
}

\noindent We found this loss to work well for our task, as it optimizes both reconstruction and discriminability.
% \vspace{-.6cm}
\section{Experiments and Analysis}
% \vspace{-.2cm}
In this section we comprehensively evaluate \modelname's capability. We first describe the experimental setup in Section~\ref{setup}. We then examine the effectiveness on unseen subject generalization in Section~\ref{unseen_subjects_evaluation}, and the decoding robustness in Section~\ref{decoding_robustness}. Next, we investigate the model’s internal representational structure via attention-based analyses in Section~\ref{attention_analysis}. Finally, we evaluate its ability to adapt to new scanner, voxel sizes, and scanning protocols on the BOLD5000 data in Section~\ref{new_scanner_adaption}. Together, these experiments provide a rigorous characterization of the model’s decoding capability, robustness, and interpretability.
\begin{figure*}[h]
    \centering
    \includegraphics[width=0.9\linewidth]{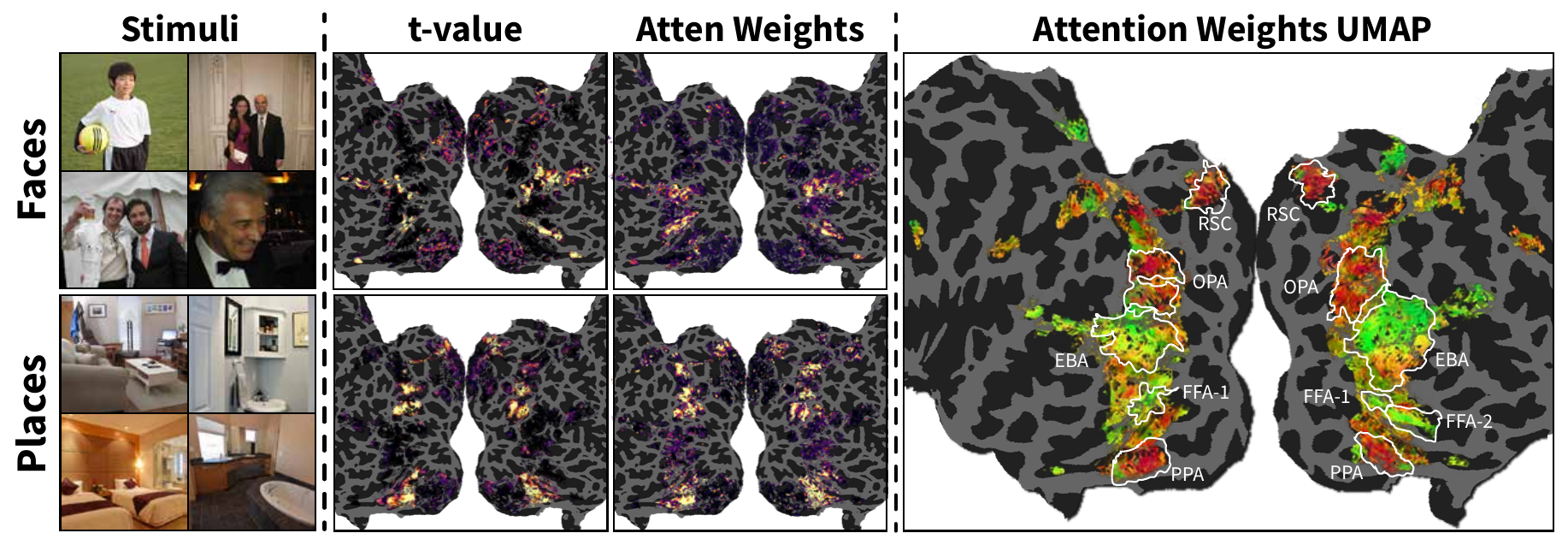}
    \vspace{-0.5em}
    \caption{
    \textbf{Semantic attention patterns in \modelname.} 
    \textbf{Left:} Example face and place stimuli used for category-specific analysis. \textbf{Middle:} Comparison of category $t$-values from an independent NSD functional localizer, with the corresponding attention-weight maps from the final self-attention layer when decoding these stimuli, showing closely matched spatial distributions. 
    \textbf{Right:} UMAP projection of voxelwise attention weights across the full test set. 
    Color-coded clusters separate body/face-selective regions in green (EBA, FFA/aTL-faces) and scene-selective regions in red (RSC, OPA, PPA).
    }
    \vspace{-10pt}
    \label{fig:8_attention_visualization}
\end{figure*}

\subsection{Experiment Setup}
\label{setup}
\noindent\textbf{Dataset.} We evaluate model performance on Natural Scenes Dataset (NSD)~\cite{allen2022massive} and further validate on BOLD5000~\cite{chang2019bold5000}. Both are large-scale fMRI datasets. NSD is the largest available 7T neural dataset, in which each subject viewed $\sim$10{,}000 images for up to three times. There is no overlap between train and test images. BOLD5000 is a 3T dataset, in which each subject viewed $\sim$5{,}000 images, but only a subset of images was viewed four times. 

For NSD, four subjects (S1, S2, S5, S7) completed the whole scanning among all eight subjects, and thus are mainly used in our experiments. For each NSD subject, roughly 9{,}000 images are uniquely seen by to that subject, while $\sim$1{,}000 images are commonly viewed by all eight subjects. To rigorously evaluate \modelname\ on novel subjects, we use the $3\times 9{,}000$ unique images from three subjects as meta-training data, the $1\times 9{,}000$ unique images from one held-out subject as the support image context, and the $1{,}000$ common images viewed by the held-out subject as the final test set. We perform analyses in subject-native volume space (\texttt{func1pt8mm}) for all NSD subjects. For the data preprocessing, voxelwise betas are $z$-scored within each session and then averaged across repeats of the same stimulus. For ROI-level evaluations, we apply a $t$-statistic threshold of $t>2$ using independent functional localizer data provided with the dataset to refine broad ROI definitions following prior work~\citep{luo2023brain}. For quantitative evaluations, we apply a voxel-quality cutoff of \texttt{ncsnr}~$>0.2$ following~\citep{conwell2024large}. For BOLD5000, we use a model trained on the four NSD subjects (no subject held-out) and evaluate directly on BOLD5000 subjects (CSI1, CSI2, CSI3) without additional training using 5-fold cross-validation. We only utilize those stimuli with four repeats and apply a cutoff of \texttt{ncsnr}$~>0.3$ as the dataset authors recommend. Voxel stimuli responses are averaged over all the repeats. 

\noindent\textbf{Training Strategy.} Our training strategy is inspired by LLM pipelines and consists of three stages: pretraining, contextual extension, and supervised fine-tuning. In the pretraining stage, we adopt an analysis-by-synthesis scheme that does not use any real fMRI data. We simulate a large population of voxels by sampling synthetic weights and corresponding beta responses with random Gaussian noise, and train the model with a fixed voxel-context size of 200. In the second stage, we introduce variable-length contexts by randomly drawing the number of voxels from $\mathrm{Uniform}(200, 4000)$, enabling the model to become robust to changes in context length. In the final fine-tuning stage, the model is optimized on real fMRI measurements, using subject-specific beta values and voxel response parameters estimated by the pretrained BraInCoRL across different image-context sizes, leading to fast convergence and effective adaptation to biologically realistic neural signals.

\noindent\textbf{Evaluation Metrics.} We evaluate cross-subject decoding on the foundational nearest-neighbor image retrieval task, which accurately reflects the capabilities of decoding models. Our method can also be extended to reconstruction tasks by incorporating an additional pretrained image generator such as IP-Adapter~\cite{ye2023ip} and Stable Diffusion~\cite{rombach2022high}. To quantitatively compare with other methods, we adopt 4 decoding quality evaluation metrics following ~\cite{kong2024toward,scotti2024mindeye2}, top-1 accuracy , top-5 accuracy, mean rank, and cosine similarity. To note \textbf{all our evaluation experiments are performed on novel subjects that are unseen by the model during training}, with exception of the no subject held-out (``no HO'') in the ablation study in Figure~\ref{fig:3_context_scaling}.

\subsection{Unseen Subject Brain Decoding}
\label{unseen_subjects_evaluation}
\noindent\textbf{SOTA Method Comparison.}
We evaluate unseen subject decoding image retrieval task with CLIP backbone following the MindEye2 protocol~\cite{scotti2024mindeye2}.  We compare against two state-of-the-art methods, MindEye2~\cite{scotti2024mindeye2} and TGBD~\cite{kong2024toward}, across all four leave-one-subject-out conditions. For fair comparisons, TGBD is retrained using its official recipe using the same dataset split as ours; MindEye2 is evaluated using its official released fine-tuned model with MNI volume anatomical alignment when inferring on novel subjects. We report the limited-context variant \textbf{using only 200 of the 9\,000 support images}, denoted \modelname-200. Quantitative and qualitative results appear in Table~\ref{tab:1_nsd_comparison} and Figure~\ref{fig:4_retrieval_visualization}, respectively. As shown above, \modelname\ delivers consistently stronger retrieval performance than both baselines on the generalizations to unseen subjects without retraining. 

\noindent\textbf{Contextual Scaling.}
We investigate how \modelname’s performance scales with the two aspects of context, image context and voxel context, respectively. The results are shown in Figure ~\ref{fig:3_context_scaling}.  A clear scaling pattern emerges across all subjects and visual backbones (CLIP, DINO, and SigLIP). Increasing either the image or voxel context size consistently improves decoding. Remarkably, with only 200 images and $4{,}000$ voxels, \modelname\ achieves similar accuracy as inference using full context (all $\sim$9{,}000 images and all higher-visual-cortex voxels). This shows that our framework requires only a fraction of subject-specific data to reach comparable decoding performance.

\noindent\textbf{Ablation Study.}
% For each held-out subject, the $\sim$9{,}000 subject-unique images serve as the \emph{image-context} support set, and the 1{,}000 images shared across all NSD subjects form the test set.
We compare four configurations, \modelname\ with synthetic data pretraining only, gradient-based functional inversion, \modelname\ trained with real data with or without subject holdout (seen subject scenario).  As illustrated in Figure~\ref{fig:3_context_scaling}, both fine-tuned variants significantly outperform the pretrained-only and direct-inversion baselines, confirming the effectiveness of \modelname. The performance gap due to subject holdout is marginal. In contrast, models trained with pretraining only or direct inversion exhibit substantially lower cosine similarity, underscoring the necessity of contextual fine-tuning for accurate cross-subject decoding.

\subsection{Robust Decoding through ROI Dropout}
\label{decoding_robustness}
We examine if \modelname\ \textbf{requires} functionally specialized cortical regions during decoding. For each semantic category (faces, places, food, and words), we first identify the test images that elicit the strongest mean beta activations within the corresponding functional voxels. We then systematically mask out the corresponding category-selective regions (e.g., removing PPA, occipital place area (OPA), and retrosplenial cortex (RSC) for scene-related stimuli) and evaluate the resulting decoding performance. As shown in Figure~\ref{fig:5_roi_exp_maskedout}, the model exhibits remarkable robustness to such targeted regional dropout. Masking category-related ROIs leads to minimal degradation for most categories, indicating that \modelname\ does not rely on any single functional region to perform aggregated decoding.

\subsection{Neural Interpretability via Attention Analysis}
\label{attention_analysis}
We analyze the internal attention dynamics of \modelname\ by extracting the attention weights from the last layer during the decoding of test images belonging to distinct semantic categories using the same activation-based selection criterion as before. As visualized in Figure~\ref{fig:8_attention_visualization}, the learned attention weights reveal highly interpretable spatial patterns. Face-related stimuli elicit elevated attention weights in voxels in the face- (FFA) and body-selective (EBA) regions, while place-related stimuli elicit elevated attention weights in place-related regions (PPA, OPA, and RSC). These results confirm that \modelname\ learns to allocate selective focus consistent with established cortical semantics.

We project the predicted voxel-wise attention weights across the entire test dataset into a three-dimensional manifold using UMAP. The resulting embedding exhibits clear semantic clustering across higher visual cortex. This emergent organization mirrors known representational gradients in visual areas, demonstrating that our model internalizes not merely \textit{how} to perform functional inversion, but \textit{where} to find semantically relevant neural representations. 
\begin{figure}[h]
  \centering
  \includegraphics[width=0.9\linewidth]{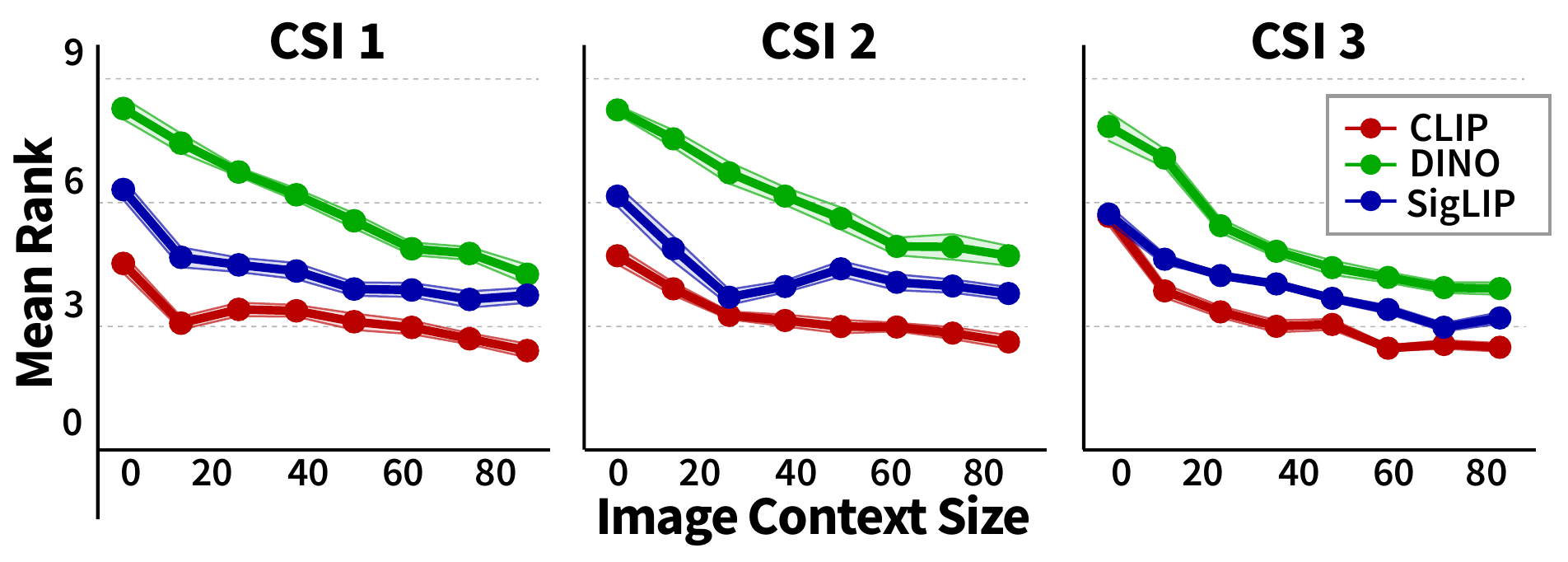}\vspace{-0.2cm}
    \caption{
    \textbf{Image-context scaling on BOLD5000.} 
    Retrieval Mean rank (lower is better) across three unseen subjects as the number of in-context image–brain pairs increases.
    }  
    \vspace{-1em}
\label{fig:6_b5k_context_scaling}
\end{figure}
\begin{figure}
    \centering
    \vspace{-0.3cm} \includegraphics[width=0.7\linewidth]{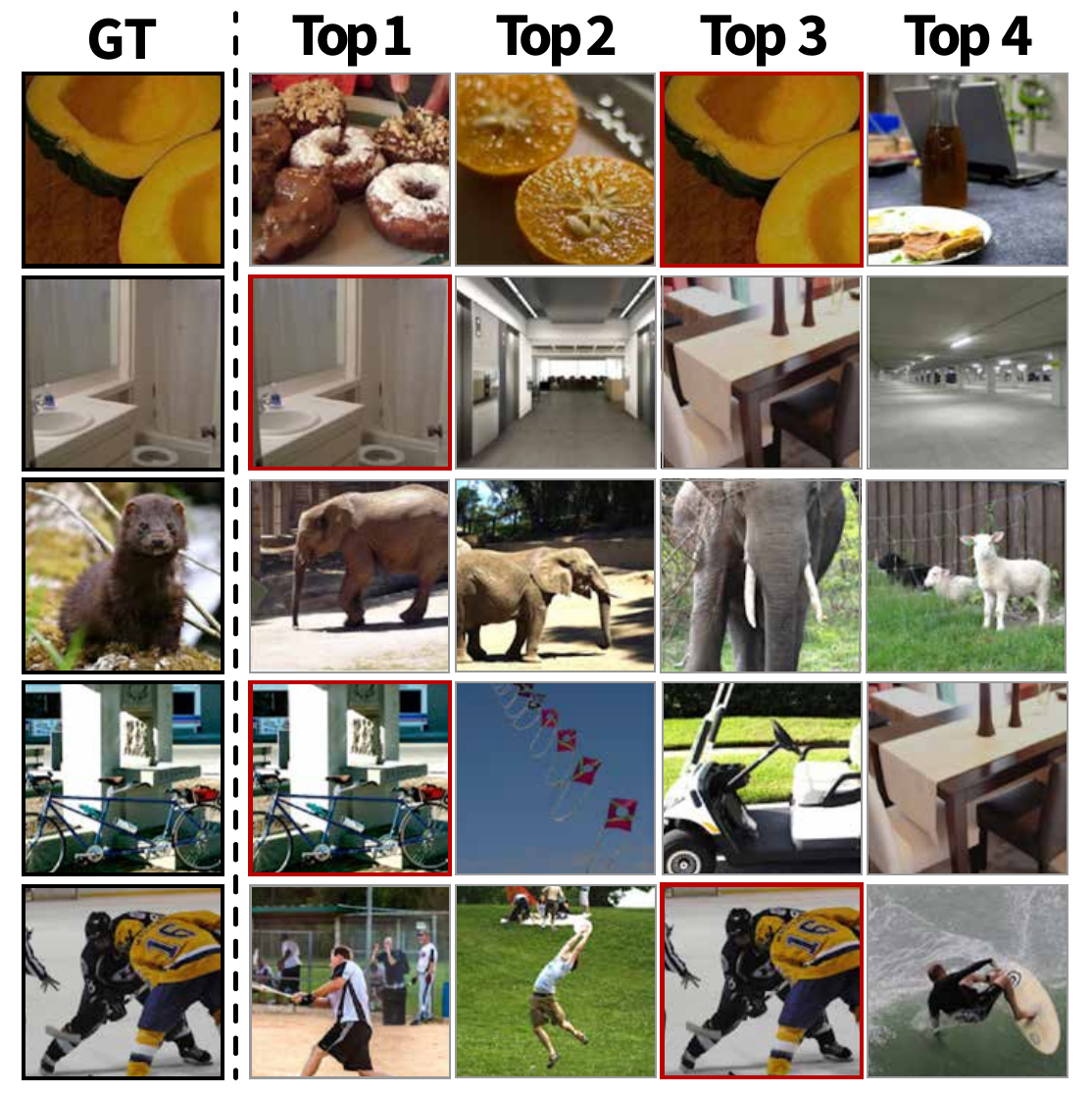}
    \vspace{-0.5em}
    \caption{
    \textbf{Top-4 image retrieval on BOLD5000.} We visualize the retrieval result on a \textbf{new-scanner unseen subject} (CSI1) using 20 images as context. The right columns display the Top-$4$ retrieved images. Red boxes indicate correct hits. Our model can generalize to novel datasets and scanning parameters without training.
    }
    \vspace{-10pt}  
    \label{fig:7_b5k_retrieval_visualization}
\end{figure}

\subsection{New Scanner Adaptation on BOLD5000}
\label{new_scanner_adaption}
We further assess cross-site generalization on BOLD5000, which differs substantially from NSD and thus provides a stringent test of new-scanner adaptation. Retrieval tasks are performed with 5-fold cross validation on BOLD5000 test images. Compared with NSD, BOLD5000 was acquired on a 3T scanner with different stimulus timing (slow event-related design with a 10\,s inter-trial interval), a substantially different image set, a different voxel size (2\,mm isotropic), and a different subject pool. Despite these shifts, \modelname\ achieves remarkable results on strong retrieval performance and exhibits a similar contextual scaling trend (Figure~\ref{fig:6_b5k_context_scaling}, Figure~\ref{fig:7_b5k_retrieval_visualization}). Results are consistent across held-out subjects and across image-encoder backbones (Table~\ref{tab:2_backbone_b5k}). Our model clearly could transfer its pretrained knowledge to new scanners, which is valuable in practical applications where retraining new models for new subjects is resource-intensive and time-consuming.
\begin{table}[h]
\centering
\caption{\textbf{Quantitative results of BOLD5000.} We directly test \modelname\ using just 20 images as in-context and a different 20 images as the test set on \textbf{three unseen subjects} from BOLD5000. Chance \texttt{Top-1 Acc} is $5\%$. All metrics are averaged across all folds of all three unseen subjects.}
\footnotesize
\setlength{\tabcolsep}{2pt}
\renewcommand{\arraystretch}{1.1}
\begin{tabular}{ccccc}
\toprule
\textbf{Backbones} & \textbf{Top-1 Acc.} & \textbf{Top-5 Acc.} & \textbf{Mean Rank} & \textbf{Cosine Sim.} \\
\midrule
CLIP         & 31.45$\pm$12.80\% & 81.67$\pm$9.42\% & 3.49$\pm$0.76 & 0.72$\pm$0.02 \\
DINOv2      & 13.99$\pm$5.83\% & 53.33$\pm$6.74\% & 6.78$\pm$0.87 & 0.08$\pm$0.01 \\
SigLIP      & 23.67$\pm$8.05\% & 73.41$\pm$8.25\% & 4.47$\pm$0.93 & 0.66$\pm$0.01 \\
\bottomrule
\end{tabular}
\label{tab:2_backbone_b5k}
\end{table}

\vspace{-.4cm}
\section{Conclusion}
% \vspace{-.2cm}
We present a foundation framework for fMRI decoding that generalizes across subjects, scanners, and acquisition protocols without any fine-tuning. By meta-learning how to invert visual encoding functions and performing hierarchical in-context inference across stimuli and voxels, BrainCoDec achieves substantial gains in data efficiency, interpretability, and cross-subject performance over strong baselines. Beyond decoding, our approach offers a principled computational lens on population-level cortical organization and demonstrates how learned functional inversion can scale across heterogeneous neural datasets. Looking forward, the same strategy can be extended to EEG, MEG, and other modalities, opening a pathway toward a universal, training-free neural decoding model for cognitive science, machine perception, and real-world BCIs.

% We introduce a foundation framework for fMRI decoding that maps neural activations to natural images and generalizes to new stimuli, subjects, scanners and protocols without finetuning. By meta-learning across subjects and leveraging in-context learning across stimuli and voxels, our approach achieves substantial gains in both data efficiency and decoding performance over strong baselines, while offering a principled lens on cortical organization in data-constrained settings. In future work, this could be extended to other neural recording modalities such as EEG, further broadening its applicability in cognitive science and BCI.
\onecolumn
% \clearpage
% \setcounter{page}{1}
% \maketitlesupplementary
\appendix
\section{Technical Appendices and Supplementary Material}
\renewcommand\thefigure{S.\arabic{figure}}    
\setcounter{figure}{0}
\renewcommand\thetable{S.\arabic{table}}   
\setcounter{table}{0}

% Here we provide further details on model architecture, implementation, results on all other NSD subjects, ablations, etc.

%%%%%%%%%%%%%%%% table of content %%%%%%%%%%%%%%%
\textbf{\Large Sections}
\begin{enumerate}
    \item Model architecture (Section~\ref{sec:model archi})
    \item Implementation details (Section~\ref{sec:implementation})
    \item More quantitative comparisons with other methods (Section~\ref{sec:quant_table})
    \item More retrieval comparisons with other methods (Section~\ref{sec:retr_vis})
    \item Context scaling of other unseen NSD subjects (Section~\ref{sec:cs_nsd_subj})
    \item Context scaling of unseen BOLD500 subjects (Section~\ref{sec:cs_b5k_subj})
    \item Attention UMAP for other NSD subjects (Section~\ref{sec:umap})
    \item More retrieval results of unseen BOLD5000 subjects (Section~\ref{sec:b5k_quant})
    \item Comparisons of model variants and ablations (Section~\ref{sec:abla})
\end{enumerate}
\clearpage
%%%%%%%%%%%%%%%% end table of content %%%%%%%%%%%%%%%

%%%%%%%%%%%%%%%% content %%%%%%%%%%%%%%%

\subsection{Model Architecture}
\label{sec:model archi}
Our \modelname\ consists of three main components:

\noindent\textbf{Voxel context token input projection.} 
For each in-context voxel, we concatenate its response function parameter $\omega_k$ and measured neural activation $\beta_k$ into a context token. We repeat this stage across voxels of interest across the brain for a single novel stimulus. A single-layer residual MLP blocks first projects this concatenated voxel context token. The residual MLP applies LayerNorm, LeakyReLU, dropout, and two linear layers with a skip connection.

\noindent\textbf{Contextual decoder transformer.} 
We employ a transformer encoder with 8 self-attention layers to perform aggregated encoder inversion across all voxel tokens and register tokens, allowing the model to infer the stimulus from encoder weights and voxel responses. Each block uses a pre-normalization architecture, we first apply LayerNorm to the inputs, scale the sequence by $logV$, where $V$ is the number of in-context voxels, and then perform self-attention. The attention output is added back with dropout. Then we apply the second LayerNorm followed by a SwiGLU feed-forward network with residual connection.

\noindent\textbf{Image embedding prediction head.}
After the transformer, we keep register tokens only, and apply an MLP to the concatenated register tokens. This yields a single predicted image embedding. 

We primarily evaluate our model using CLIP, due to its excellent visual brain predictivity~\citep{conwell2024large}, and additionally assess variants based on DINOv2~\citep{oquab2023dinov2} and SigLIP~\citep{zhai2023sigmoid}. The CLIP variant (encoding dimension $E = 512$) contains approximately 55.70M parameters, while the DINOv2 ($E = 768$) and SigLIP ($E = 1152$) variants comprise roughly 88.76M and 157.35M parameters, respectively. For all models we utilize the ViT-B variant.

\vspace{2cm}
% \clearpage

%%%%%%%%%%%%%%%%%%%%

\subsection{Implementation Details}
\label{sec:implementation}
Training is implemented in PyTorch on two NVIDIA RTX 4090 GPUs (48GB each). At each training step, we sample a batch of in-context voxel tokens together with their target image-embedding vectors and feed them through \modelname\ to obtain predicted embeddings. We train the model with a supervised objective that combines a cosine-similarity loss and an InfoNCE loss between predicted and ground-truth embeddings. Dropout is applied in all residual and attention blocks to regularize the model and mitigate overfitting. We optimize \modelname\ using AdamW with an initial learning rate of $1\times 10^{-5}$ and a decoupled weight decay of $1\times 10^{-2}$. In the first pretraining stage, each mini-batch samples a fixed set of 200 in-context voxels. In the second context-extension stage and the third finetuning stage, each mini-batch randomly samples between 200 and 4000 in-context voxels. The learning rate is scheduled with a cosine-annealing scheduler over the total number of training steps, gradually decaying to a minimum of $1\times 10^{-6}$. We use the HuggingFace \texttt{Accelerate} library to jointly prepare the model, optimizer, data loaders, and scheduler for (potentially) distributed training. The same training protocol is applied to the CLIP, DINOv2, and SigLIP variants, differing only in the choice of backbone embedding dimension.

In the main paper, we focus on NSD S1/S2/S5/S7, as these are the four subjects that completed scanning from the dataset. We train $15$ models total based on three backbones. For each backbone we train five variants (four where a single subject is held out, and one model where we train on all four subjects). Note, all of these models are effectively fine-tuned variants of the model that was trained with synthetic data only. The variants where a single subject is held out is used respectively for testing on S1/S2/S5/S7 from NSD to ensure there is no data contamination. For NSD S3/S4/S6/S8 and BOLD5000, we use the variant trained on all four NSD complete subject.

Our code will be open sourced once the review process is concluded. We thank the reviewers for your understanding.

For this supplemental, we first present the results for the subjects that completed NSD scanning (S1/S2/S5/S7), then we present the subjects that did not (S3/S4/S6/S8). Unless otherwise noted, in all cases the model has not seen data from a particular subject during training.
\clearpage

% SOTA comparison
\subsection{Quantitative table for S2-8}
\label{sec:quant_table}
\begin{table}[h]
\centering
\caption{\textbf{Quantitative comparison on NSD Subjects 1, 2, 5, and 7.}}
\renewcommand{\arraystretch}{1.1}
\setlength{\tabcolsep}{8pt} % Adjust padding between columns
\begin{tabular}{lcccc}
\toprule
\textbf{Model} & \textbf{S1} & \textbf{S2} & \textbf{S5} & \textbf{S7} \\

\midrule
\multicolumn{5}{c}{$\%$ \textbf{Top-1 Accuracy ($\uparrow$)}} \\
\cmidrule(lr){1-5}
MindEye2       & $4.11 \pm 1.41$& $3.82 \pm 1.10$& $2.87\pm 1.19$& $2.51\pm 1.64$\\
TGBD           & $1.27 \pm 0.16$& $0.56 \pm 0.12$& $0.84 \pm 0.16$& $0.39 \pm 0.09$\\
\textbf{BrainCodec-200} & $\mathbf{25.5} \pm \mathbf{3.02}$& $\mathbf{22.9} \pm \mathbf{2.98}$& $\mathbf{23.2} \pm \mathbf{2.63}$& $\mathbf{19.2} \pm \mathbf{2.42}$\\

\midrule
\multicolumn{5}{c}{$\%$ \textbf{Top-5 Accuracy ($\uparrow$)}} \\
\cmidrule(lr){1-5}
MindEye2       & $12.9 \pm 2.55$& $10.7 \pm 3.14$& $9.58\pm 3.61$& $6.49 \pm 2.87$\\
TGBD           & $3.89 \pm 1.25$& $2.33 \pm 0.91$& $3.34 \pm 0.99$& $1.41 \pm 0.78$\\
\textbf{BrainCodec-200} & $\mathbf{56.6} \pm \mathbf{3.21}$& $\mathbf{52.4} \pm \mathbf{4.08}$& $\mathbf{55.8} \pm \mathbf{2.47}$& $\mathbf{51.2} \pm \mathbf{3.50}$\\

% \midrule
% \multicolumn{5}{c}{$\%$ \textbf{Mean Rank ($\downarrow$)}} \\
% \cmidrule(lr){1-5}
% MindEye2       & $74.1 \pm 6.2$& $75.3 \pm 7.2$& $78.1 \pm 9.42$& $76.9 \pm 8.0$\\
% TGBD           & $145.5 \pm 8.6$& $152.6 \pm 9.4$& $141.4 \pm 9.6$& $148.4 \pm 7.3$\\
% \textbf{BrainCodec-200} & $\mathbf{13.3} \pm \mathbf{1.4}$& $\mathbf{12.7} \pm \mathbf{1.0}$& $\mathbf{11.8} \pm \mathbf{0.8}$& $\mathbf{11.2} \pm \mathbf{0.9}$\\

\midrule
\multicolumn{5}{c}{$\%$ \textbf{Mean Rank ($\downarrow$)}} \\
\cmidrule(lr){1-5}
MindEye2       & $24.70 \pm 2.07$& $25.10 \pm 2.40$& $26.03 \pm 3.14$& $25.63 \pm 2.67$\\
TGBD           & $48.50 \pm 2.87$& $50.87 \pm 3.13$& $47.13 \pm 3.20$& $49.47 \pm 2.43$\\
\textbf{BrainCodec-200} & $\mathbf{4.43} \pm \mathbf{0.47}$& $\mathbf{4.23} \pm \mathbf{0.33}$& $\mathbf{3.93} \pm \mathbf{0.27}$& $\mathbf{3.73} \pm \mathbf{0.30}$\\

% \midrule
% \multicolumn{5}{c}{\textbf{Cosine Similarity ($\uparrow$)}} \\
% \cmidrule(lr){1-5}
% MindEye2       & $0.11 \pm 0.04$& $0.11 \pm 0.03$& $0.12 \pm 0.05$& $0.12 \pm 0.04$\\
% TGBD           & $0.06 \pm 0.01$& $0.06 \pm 0.01$& $0.05 \pm 0.01$& $0.04 \pm 0.01$\\
% \textbf{BrainCodec-200} & $\mathbf{0.81} \pm \mathbf{0.02}$& $\mathbf{0.80} \pm \mathbf{0.02}$& $\mathbf{0.79} \pm \mathbf{0.03}$& $\mathbf{0.79} \pm \mathbf{0.02}$\\
\bottomrule
\end{tabular}
\end{table}

\begin{table}[h]
\centering
\caption{\textbf{Quantitative comparison on NSD Subjects 3, 4, 6, and 8.}}
\renewcommand{\arraystretch}{1.1}
\setlength{\tabcolsep}{8pt}
\begin{tabular}{lcccc}
\toprule
\textbf{Model} & \textbf{S3} & \textbf{S4} & \textbf{S6} & \textbf{S8} \\
\midrule
\multicolumn{5}{c}{$\%$ \textbf{Top-1 Accuracy ($\uparrow$)}} \\
\cmidrule(lr){1-5}
MindEye2       & $3.50 \pm 1.13$ & $3.19 \pm 1.16$ & $2.69 \pm 1.42$ & $2.33 \pm 1.86$ \\
TGBD           & $0.65 \pm 0.13$ & $0.75 \pm 0.15$ & $0.61 \pm 0.12$ & $0.17 \pm 0.05$ \\
\textbf{BrainCodec-200} & $\mathbf{19.0} \pm \mathbf{1.86}$& $\mathbf{16.1} \pm \mathbf{1.75}$& $\mathbf{20.1} \pm \mathbf{2.52}$& $\mathbf{14.4} \pm \mathbf{1.56}$\\
\midrule
\multicolumn{5}{c}{$\%$ \textbf{Top-5 Accuracy ($\uparrow$)}} \\
\cmidrule(lr){1-5}
MindEye2       & $10.33 \pm 3.30$ & $9.95 \pm 3.45$ & $8.04 \pm 3.24$ & $4.95 \pm 2.50$ \\
TGBD           & $2.67 \pm 0.94$  & $3.00 \pm 0.96$ & $2.38 \pm 0.89$ & $0.44 \pm 0.68$ \\
\textbf{BrainCodec-200} & $\mathbf{48.3} \pm \mathbf{2.34}$& $\mathbf{42.3} \pm \mathbf{3.01}$& $\mathbf{48.7} \pm \mathbf{3.00}$& $\mathbf{53.3} \pm \mathbf{4.02}$\\

\midrule
\multicolumn{5}{c}{$\%$ \textbf{Mean Rank ($\downarrow$)}} \\
\cmidrule(lr){1-5}
MindEye2               & $25.40 \pm 2.63$& $25.73 \pm 2.90$& $25.83 \pm 2.90$& $25.43 \pm 2.43$\\
TGBD                   & $49.63 \pm 3.17$& $48.37 \pm 3.17$& $48.30 \pm 2.80$& $50.63 \pm 2.07$\\
\textbf{BrainCodec-200}& $\mathbf{4.97} \pm \mathbf{0.30}$& $\mathbf{5.97} \pm \mathbf{0.30}$& $\mathbf{4.53} \pm \mathbf{0.30}$& $\mathbf{3.03} \pm \mathbf{0.27}$\\

% \midrule
% \multicolumn{5}{c}{\textbf{Mean Rank ($\downarrow$)}} \\
% \cmidrule(lr){1-5}
% MindEye2       & $76.2 \pm 7.9$ & $77.2 \pm 8.7$ & $77.5 \pm 8.7$ & $76.3 \pm 7.3$ \\
% TGBD           & $148.9 \pm 9.5$ & $145.1 \pm 9.5$ & $144.9 \pm 8.4$ & $151.9 \pm 6.2$ \\
% \textbf{BrainCodec-200} & $\mathbf{14.9} \pm \mathbf{0.9}$& $\mathbf{17.9} \pm \mathbf{0.9}$& $\mathbf{13.6} \pm \mathbf{0.9}$& $\mathbf{9.1} \pm \mathbf{0.8}$\\

% \midrule
% \multicolumn{5}{c}{\textbf{Cosine Similarity ($\uparrow$)}} \\
% \cmidrule(lr){1-5}
% MindEye2       & $0.11 \pm 0.04$ & $0.12 \pm 0.04$ & $0.12 \pm 0.04$ & $0.12 \pm 0.05$ \\
% TGBD           & $0.06 \pm 0.01$ & $0.05 \pm 0.01$ & $0.04 \pm 0.01$ & $0.04 \pm 0.01$ \\
% \textbf{BrainCodec-200} & $\mathbf{0.78} \pm \mathbf{0.03}$& $\mathbf{0.77} \pm \mathbf{0.03}$& $\mathbf{0.79} \pm \mathbf{0.03}$& $\mathbf{0.77} \pm \mathbf{0.02}$\\
\bottomrule
\end{tabular}
\end{table}
\clearpage

% \newgeometry{top=0.6in}
\subsection{Retrieval visualizations for NSD}
\label{sec:retr_vis}
\begin{figure}[h]
\centering
\vspace{-10pt}
\begin{minipage}[b]{0.67\textwidth}
    \centering
    \includegraphics[width=\linewidth]{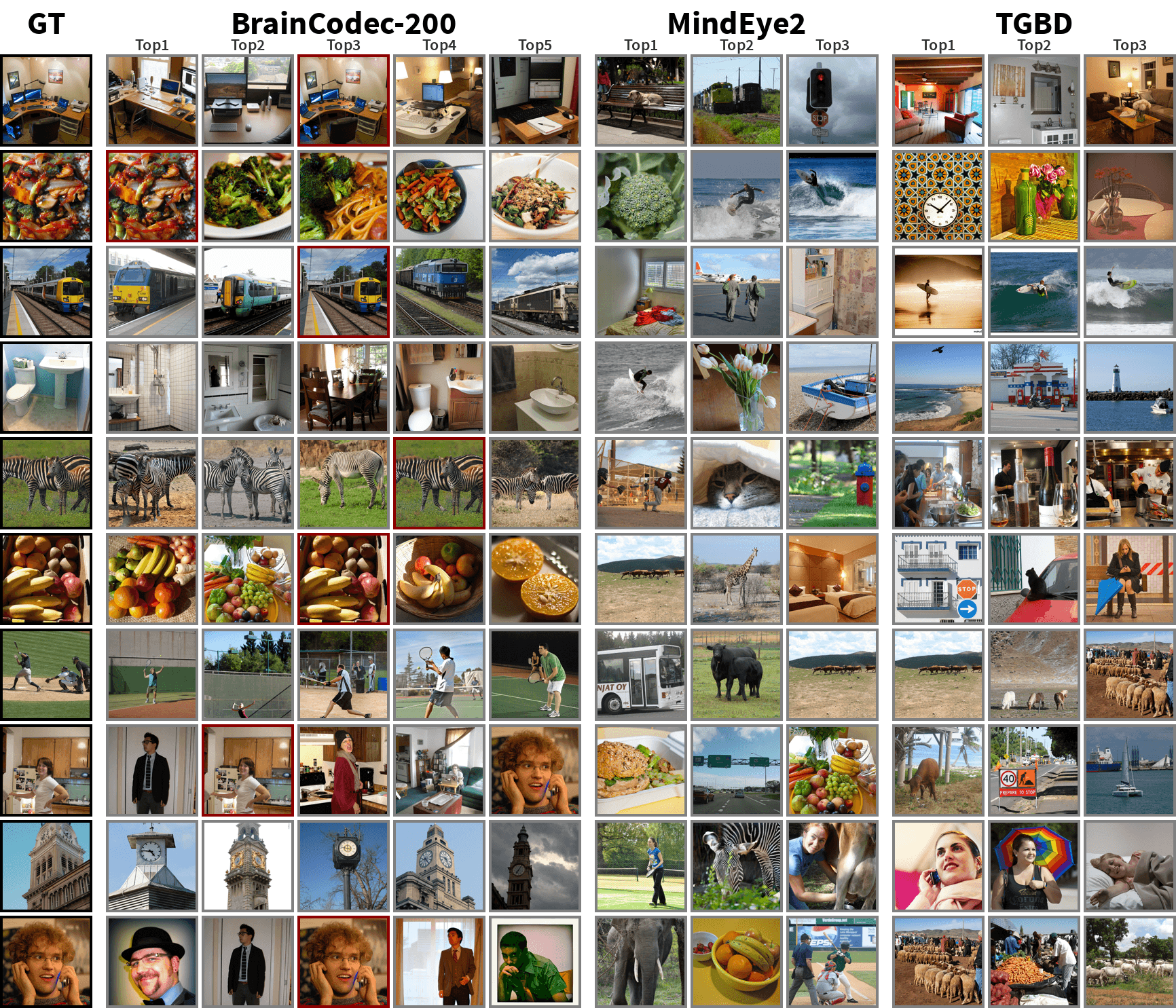}
    \captionof{figure}{Image retrieval comparison on an unseen subject (S1).}
\end{minipage}
\begin{minipage}[b]{0.67\textwidth}
    \centering
    \includegraphics[width=\linewidth]{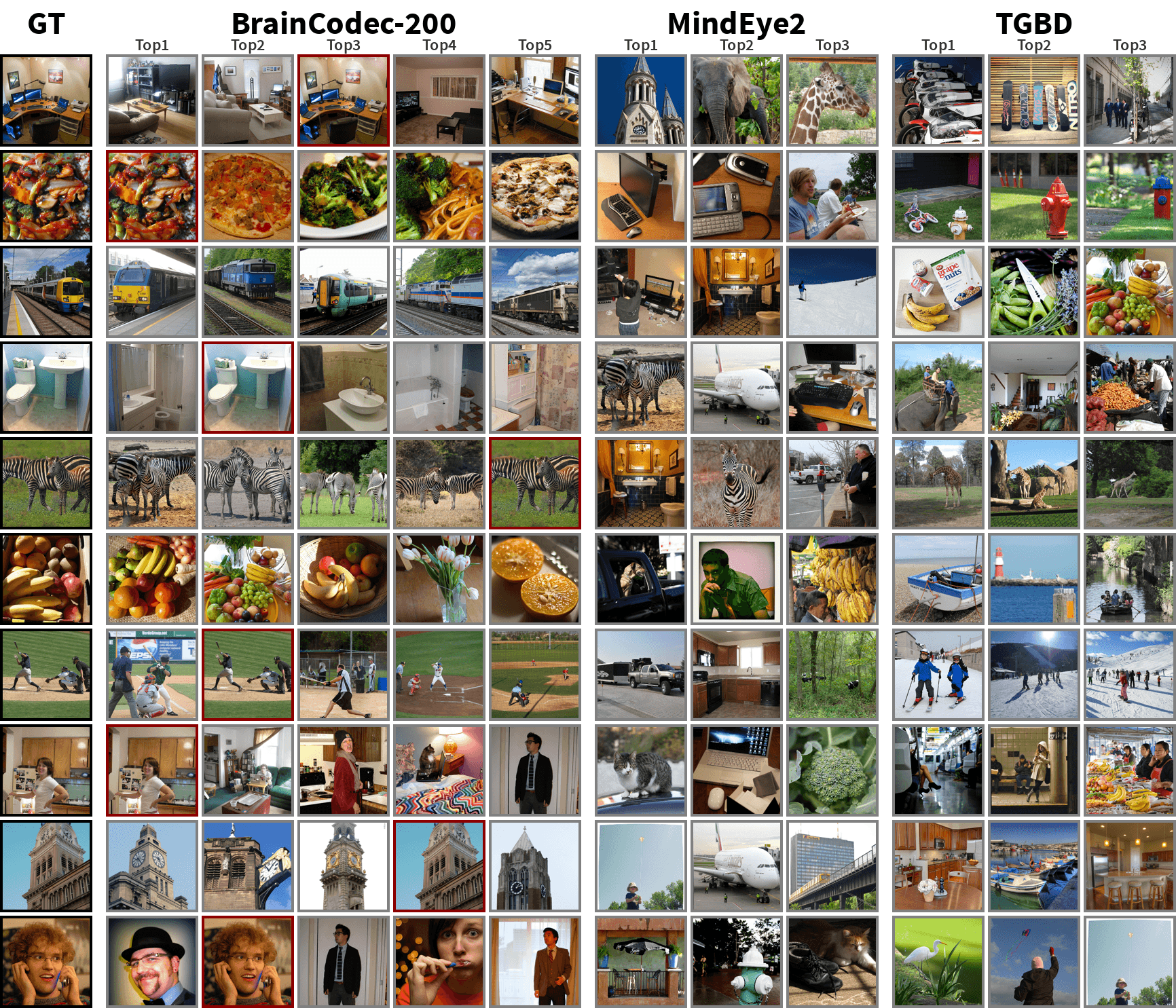}
    \captionof{figure}{Image retrieval comparison on an unseen subject (S2).}
\end{minipage}
\end{figure}

\begin{figure}[h]
\centering
\begin{minipage}[b]{0.67\textwidth}
    \centering
    \includegraphics[width=\linewidth]{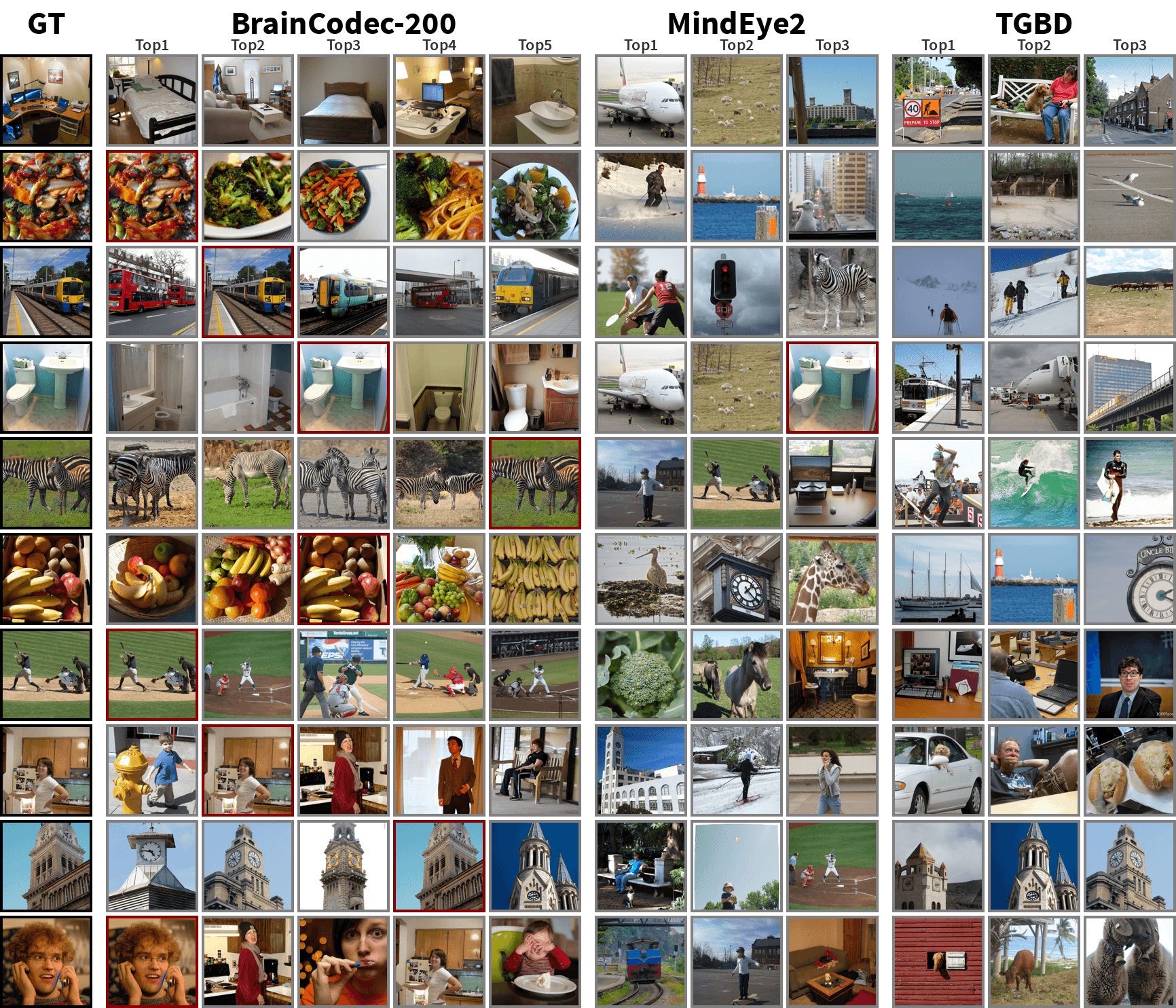}
    \captionof{figure}{Image retrieval comparison on an unseen subject (S5).}
\end{minipage}
\begin{minipage}[b]{0.67\textwidth}
    \centering
    \includegraphics[width=\linewidth]{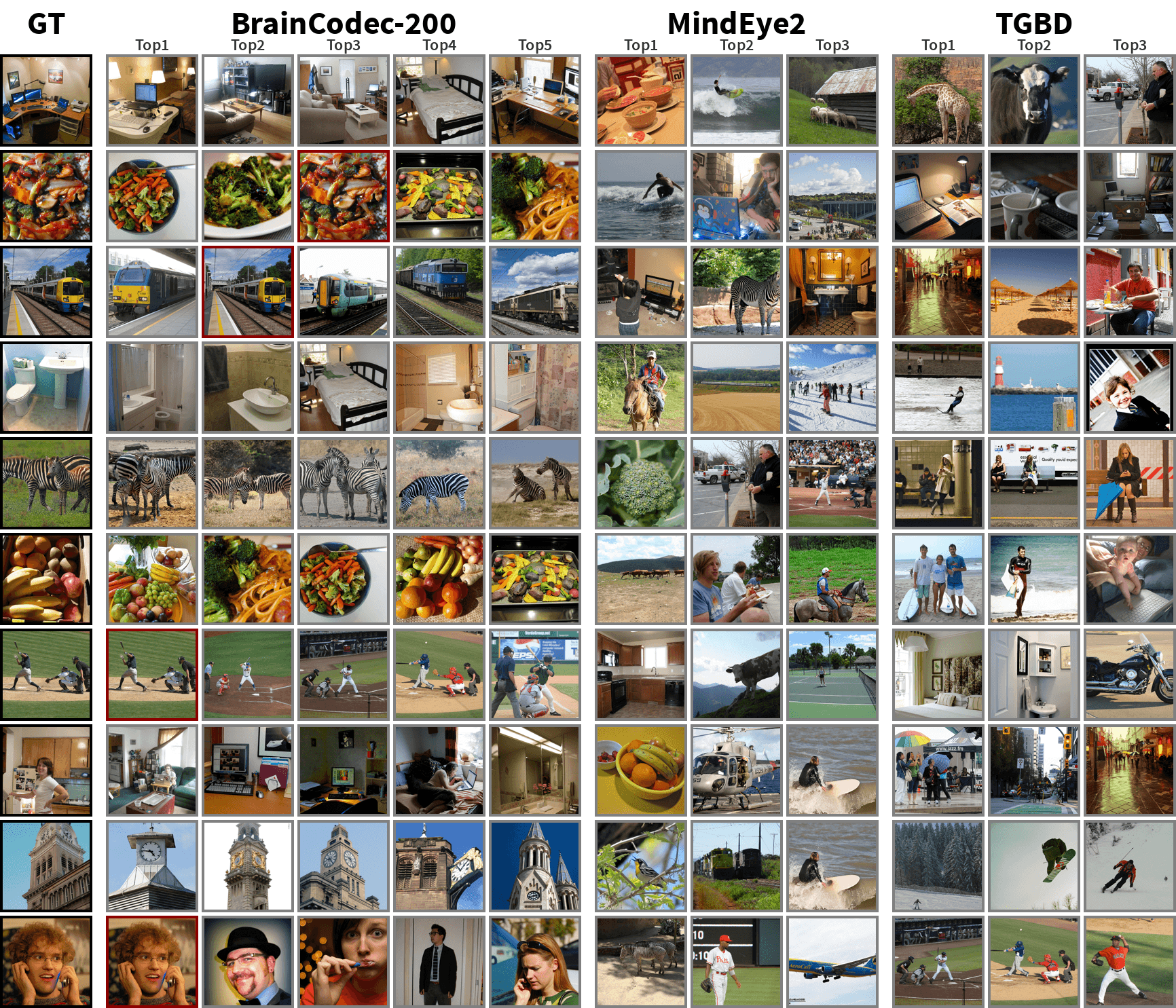}
    \captionof{figure}{Image retrieval comparison on an unseen subject (S7).}
\end{minipage}
\end{figure}

\begin{figure}[h]
\centering
\begin{minipage}[b]{0.67\textwidth}
    \centering
    \includegraphics[width=\linewidth]{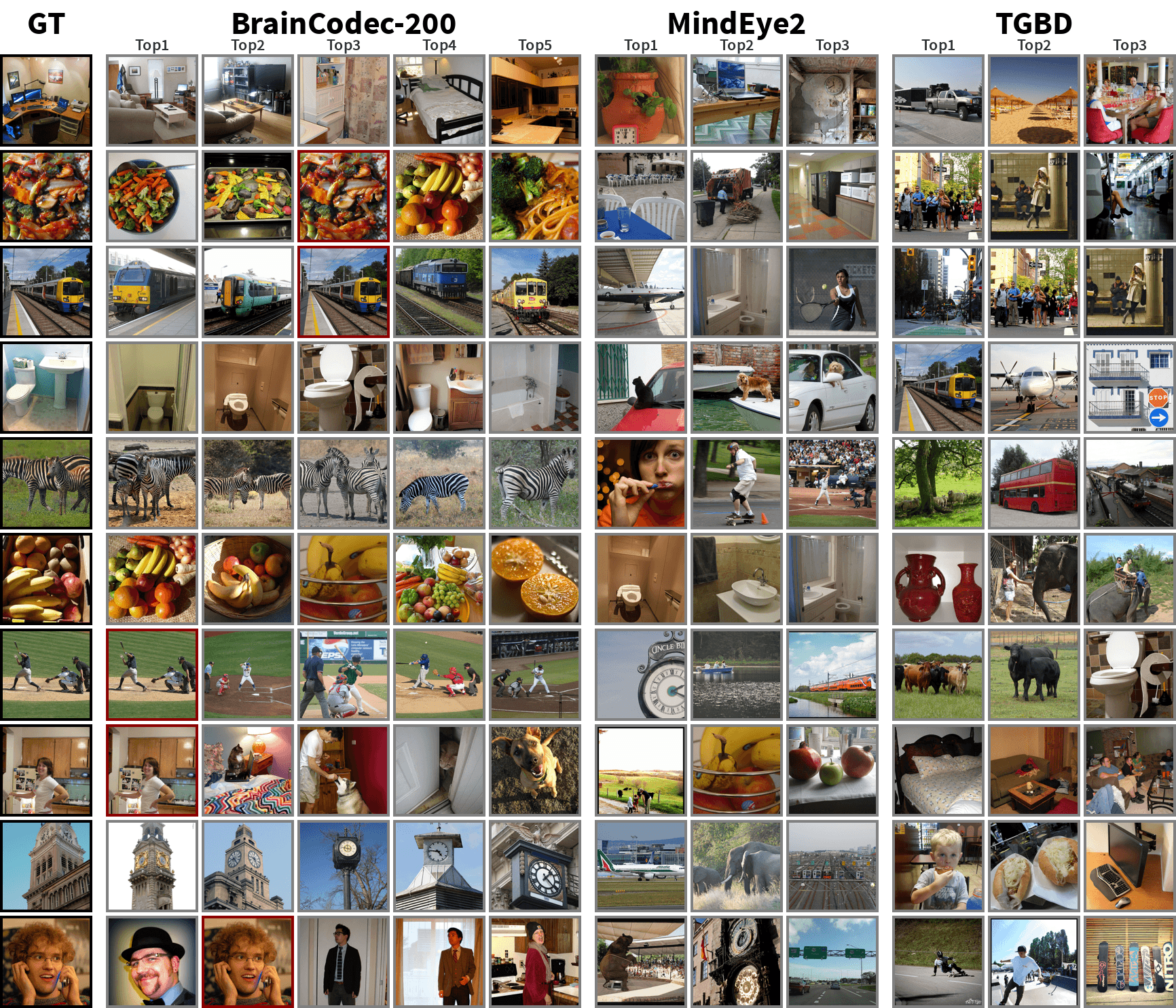}
    \captionof{figure}{Image retrieval comparison on an unseen subject (S3).}
\end{minipage}
\begin{minipage}[b]{0.67\textwidth}
    \centering
    \includegraphics[width=\linewidth]{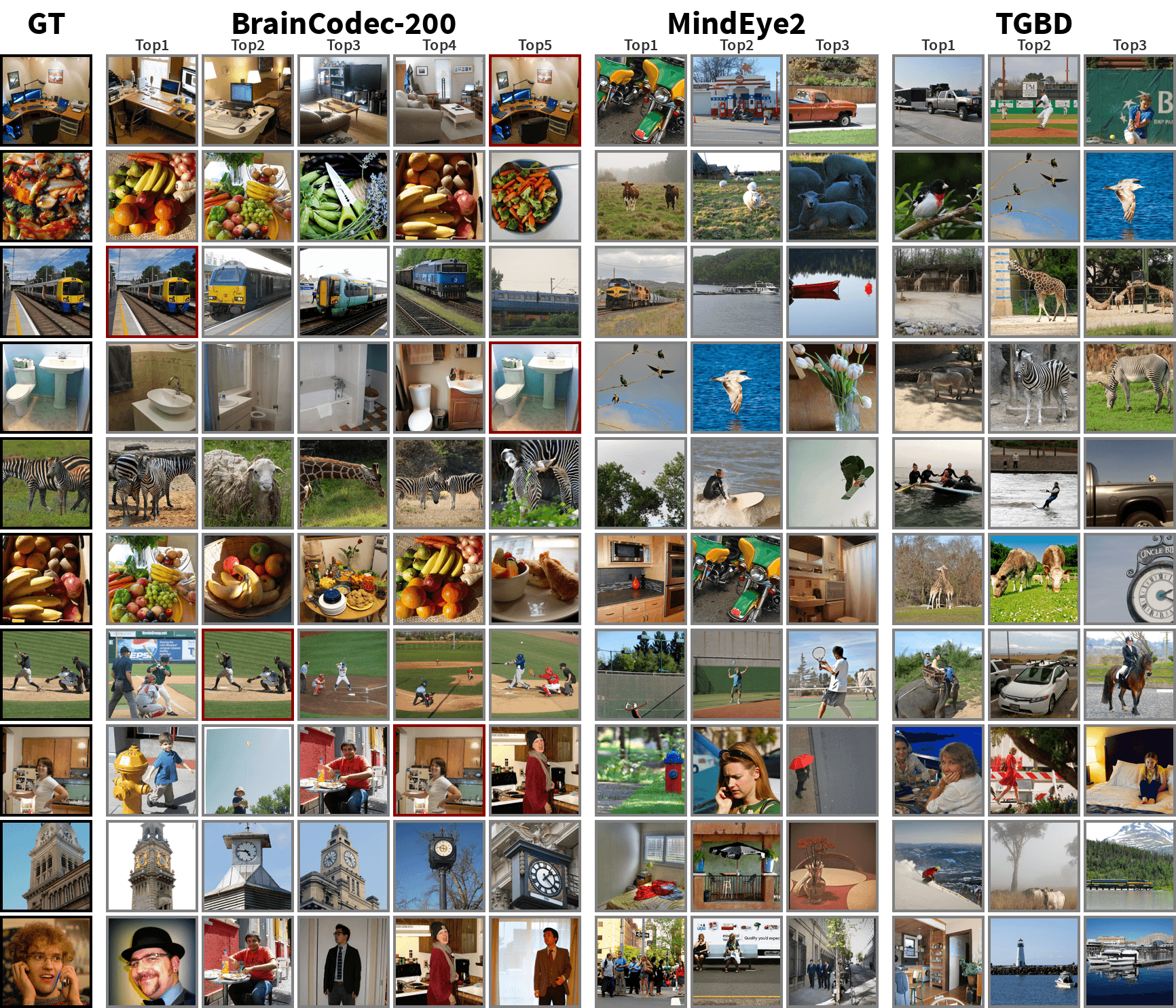}
    \captionof{figure}{Image retrieval comparison on an unseen subject (S4).}
\end{minipage}
\end{figure}

\begin{figure}[h]
\centering
\begin{minipage}[b]{0.67\textwidth}
    \centering
    \includegraphics[width=\linewidth]{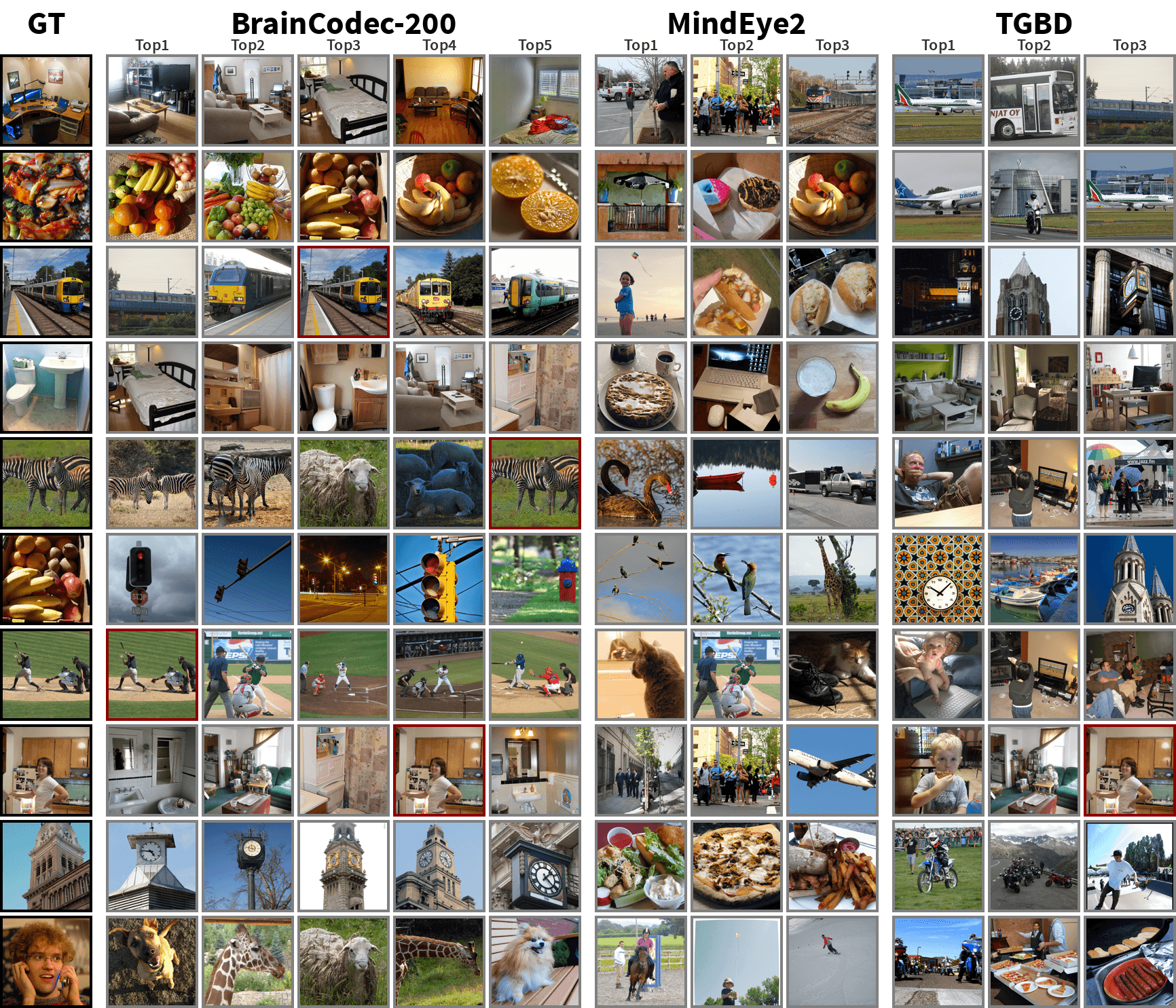}
    \captionof{figure}{Image retrieval comparison on an unseen subject (S6).}
\end{minipage}
\begin{minipage}[b]{0.67\textwidth}
    \centering
    \includegraphics[width=\linewidth]{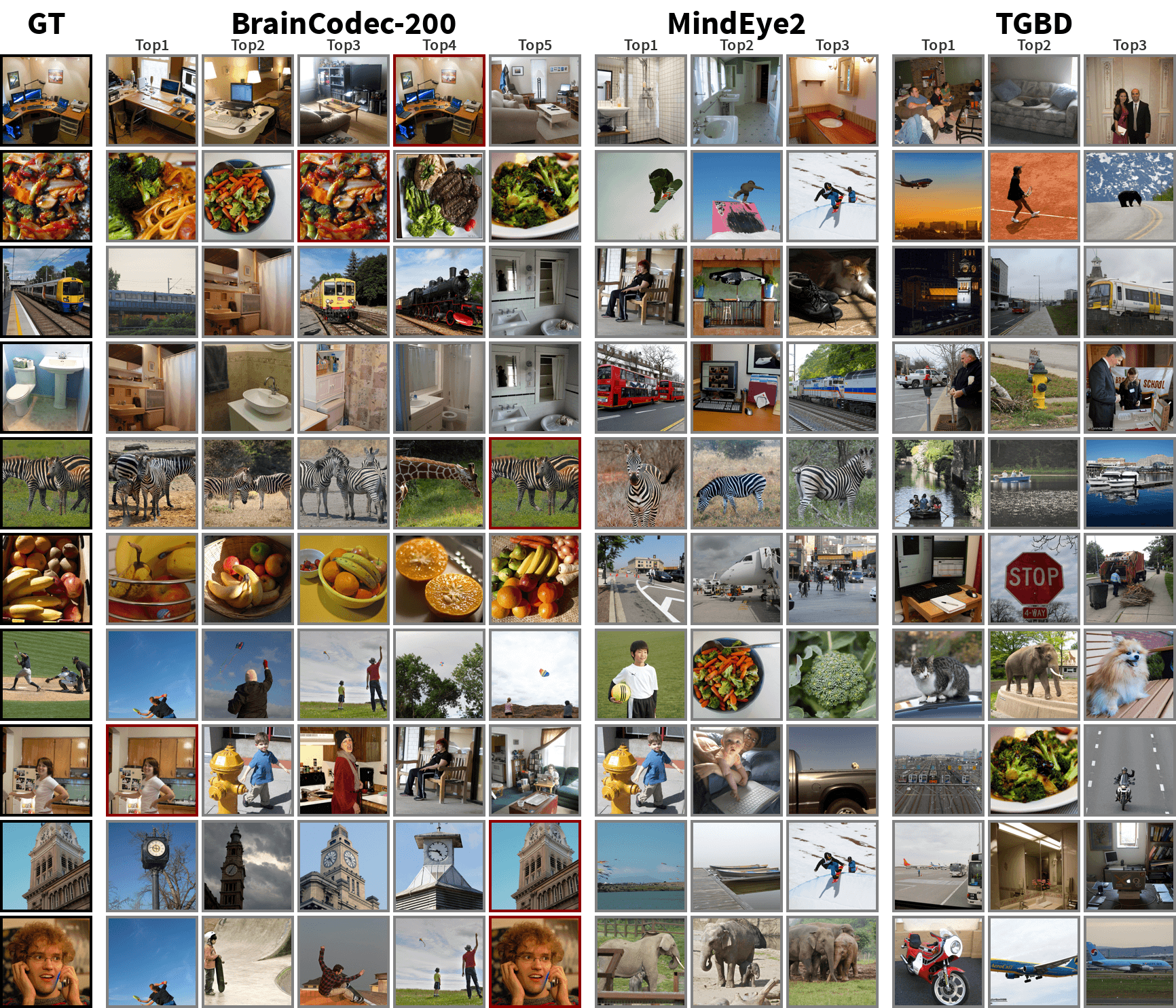}
    \captionof{figure}{Image retrieval comparison on an unseen subject (S8).}
\end{minipage}
\end{figure}

% \begin{figure*}[t]
%     \centering
%     \begin{subfigure}[b]{0.78\textwidth}
%         \centering
%         \includegraphics[width=\linewidth]{suppl_fig_png/retrieval_vis_comparison_subj1.png}
%         \caption{Image retrieval comparison on an unseen subject (S1).}
%         \label{fig:retrieval_s1}
%     \end{subfigure}
%     \begin{subfigure}[b]{0.78\textwidth}
%         \centering
%         \includegraphics[width=\linewidth]{suppl_fig_png/retrieval_vis_comparison_subj2.png}
%         \caption{Image retrieval comparison on an unseen subject (S2).}
%         \label{fig:retrieval_s2}
%     \end{subfigure}
%     % \caption{Image retrieval comparison on subjects unseen during training.}
%     \label{fig:retrieval_page1}
% \end{figure*}
% \clearpage
% \begin{figure*}[t]
%     \centering
%     \begin{subfigure}[b]{0.78\textwidth}
%         \centering
%         \includegraphics[width=\linewidth]{suppl_fig_png/retrieval_vis_comparison_subj5.png}
%         \caption{Image retrieval comparison on an unseen subject (S5).}
%         \label{fig:retrieval_s5}
%     \end{subfigure}
%     \begin{subfigure}[b]{0.78\textwidth}
%         \centering
%         \includegraphics[width=\linewidth]{suppl_fig_png/retrieval_vis_comparison_subj7.png}
%         \caption{Image retrieval comparison on an unseen subject (S7).}
%         \label{fig:retrieval_s7}
%     \end{subfigure}
%     % \caption{Image retrieval comparison on subjects unseen during training.}
%     \label{fig:retrieval_page2}
% \end{figure*}

% \restoregeometry
\clearpage

\subsection{Context scaling of other unseen NSD subjects}
\label{sec:cs_nsd_subj}
\begin{figure}[htbp]
    \centering
    \includegraphics[width=\linewidth]{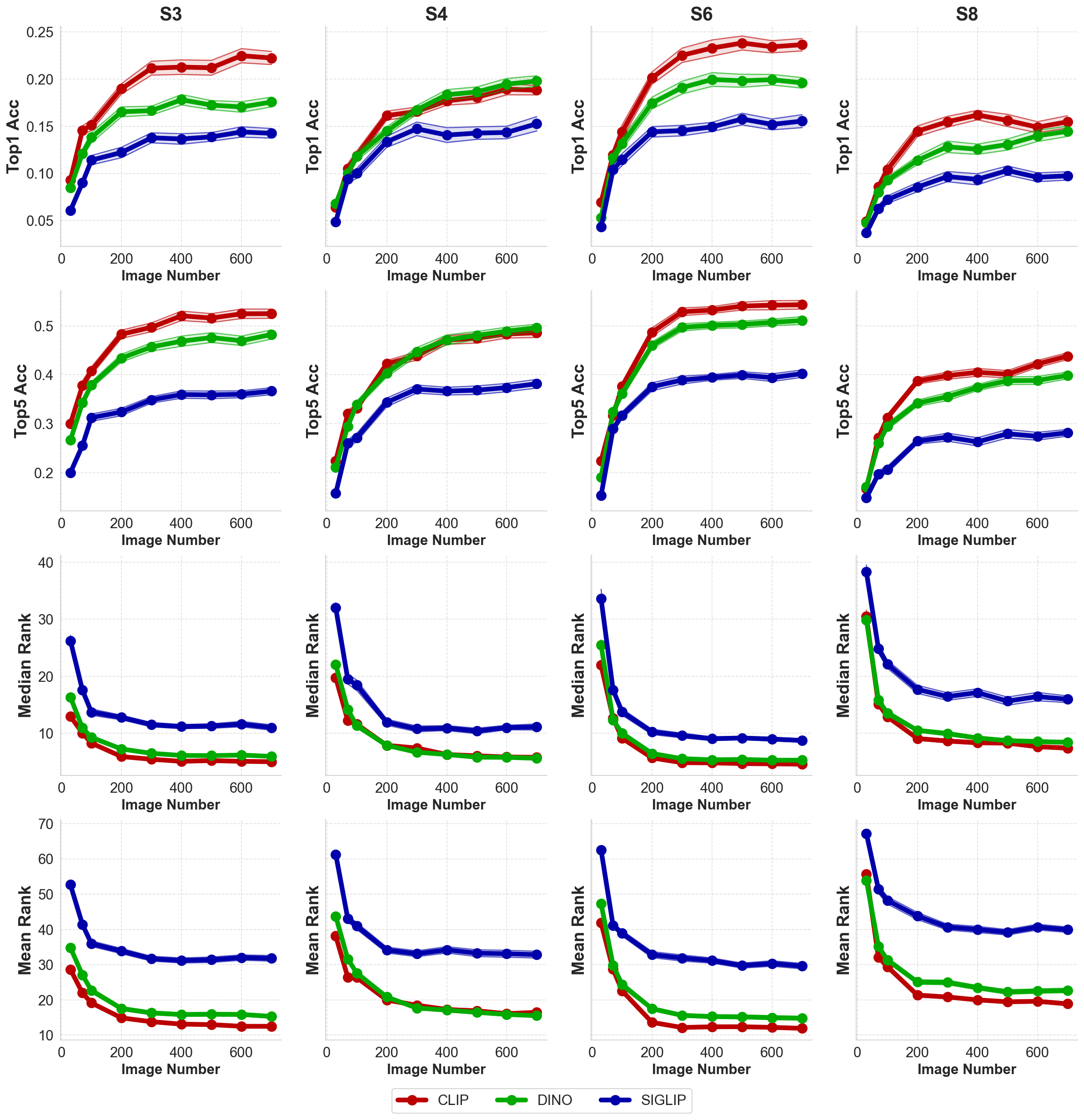}
    \caption{Image-context scaling of \modelname\ on NSD subjects 3, 4, 6, and 8.}
\end{figure}

\begin{figure}[h]
    \centering
    \includegraphics[width=\linewidth]{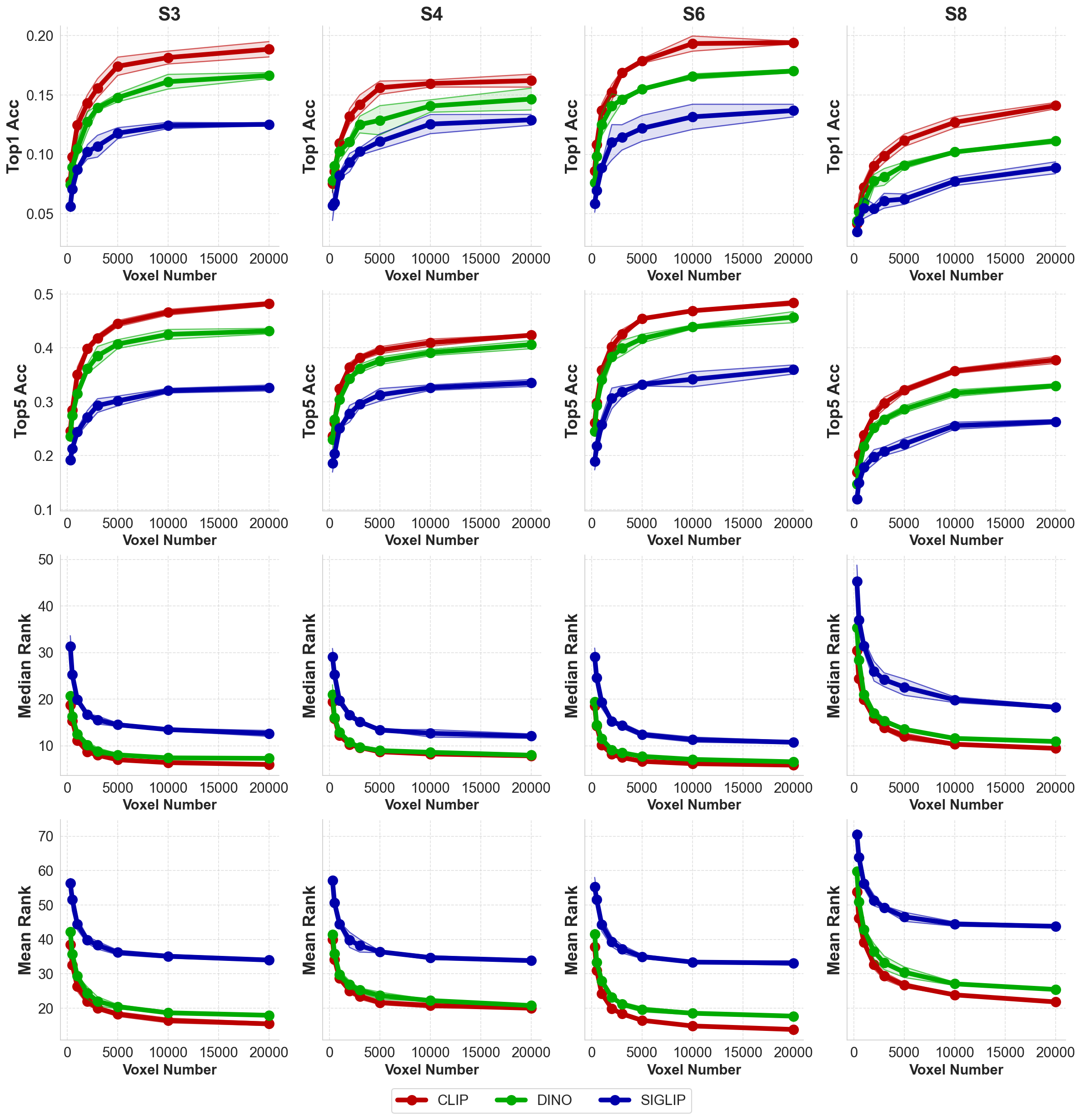}
    \caption{Voxel-context scaling of~\modelname\ on NSD subjects 3, 4, 6, and 8.}
\end{figure}

\clearpage

\subsection{Context scaling of unseen BOLD500 subjects}
\label{sec:cs_b5k_subj}
\begin{figure}[h]
    \centering
    \includegraphics[width=\linewidth]{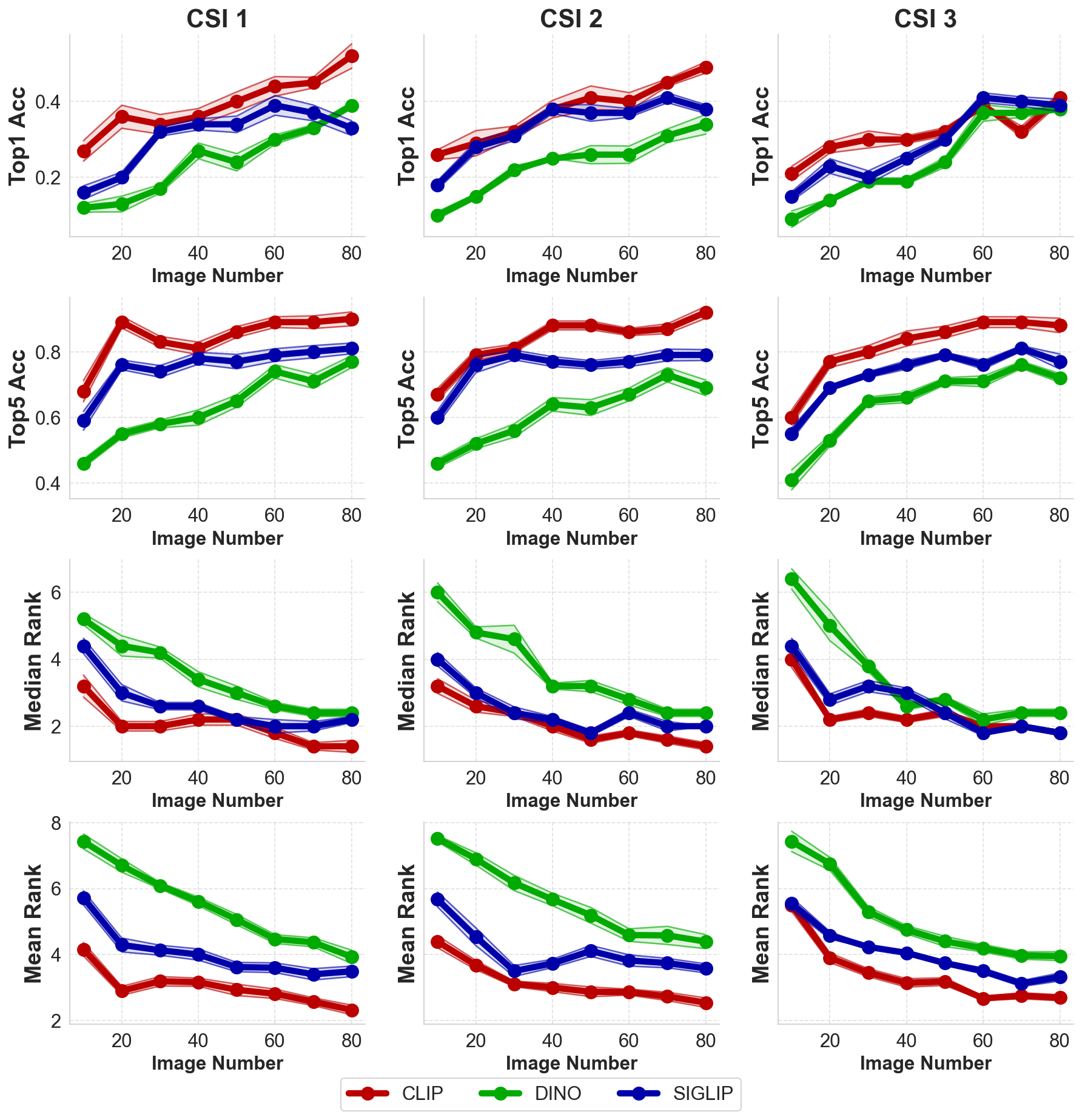}
    \caption{Image-context scaling of \modelname\ on BOLD5000 subjects.}
\end{figure}

\begin{figure}[h]
    \centering
    \includegraphics[width=\linewidth]{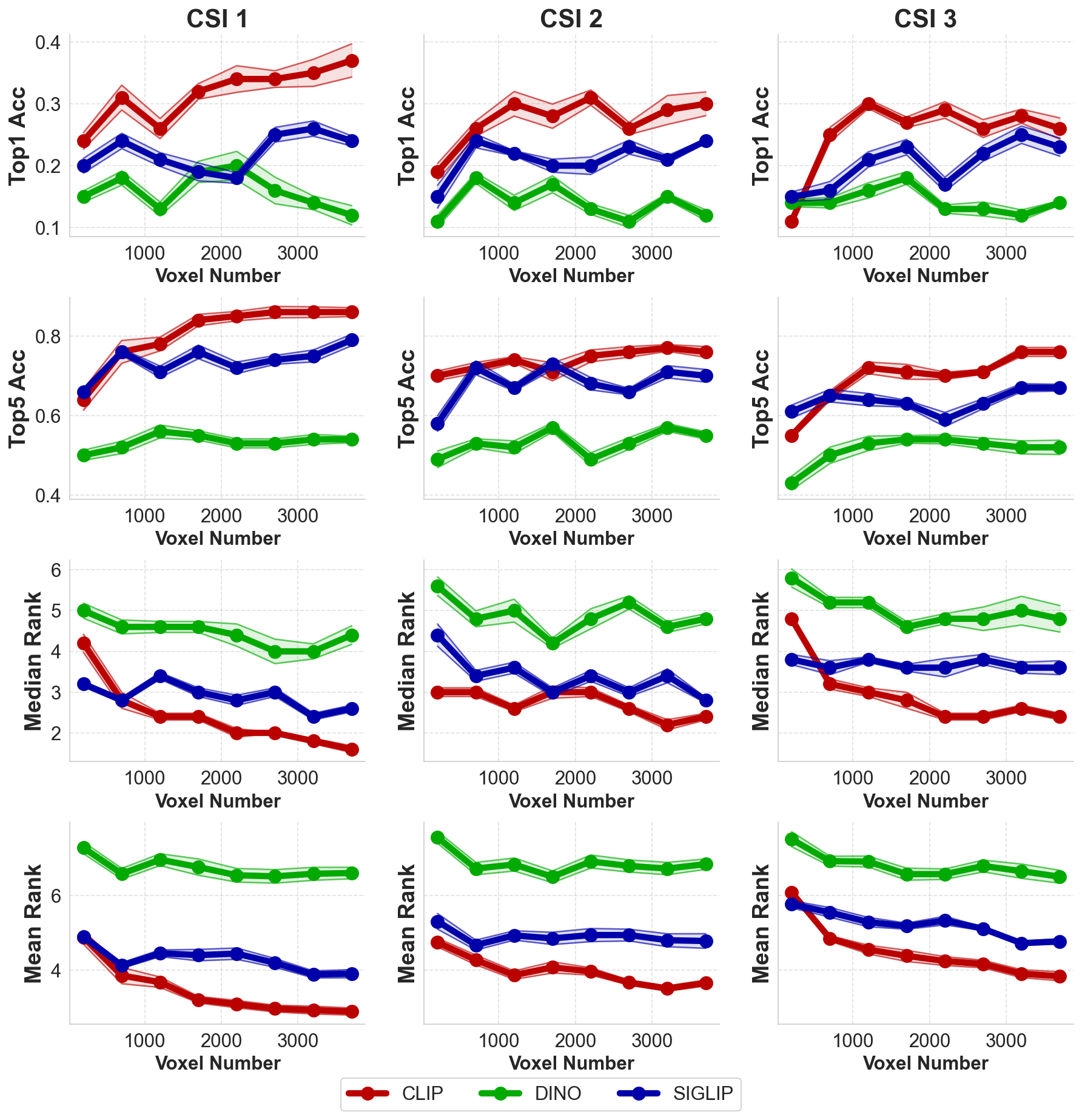}
    \caption{Voxel-context scaling of \modelname\ on BOLD5000 subjects.}
\end{figure}

\clearpage

% Attention Umap
\subsection{Attention UMAP for other subjects}
\label{sec:umap}
\begin{figure}[h]
    \centering
    \includegraphics[width=0.8\linewidth]{suppl_fig_png/attenmap_257.png}
    \caption{Semantic attention patterns in \modelname.}
\end{figure}

\clearpage

% \subsection{Quantitative results on BOLD5000 CSI 1 2 3}
% \input{suppl_table/metrics_b5k}
% \clearpage

\subsection{More Retrieval Results on unseen BOLD5000 Subjects}
\label{sec:b5k_quant}
\thispagestyle{empty}
\begin{figure}[h]
\centering
\vspace{-16pt}
\begin{subfigure}{0.28\textwidth}
    \centering
    \includegraphics[width=\linewidth]{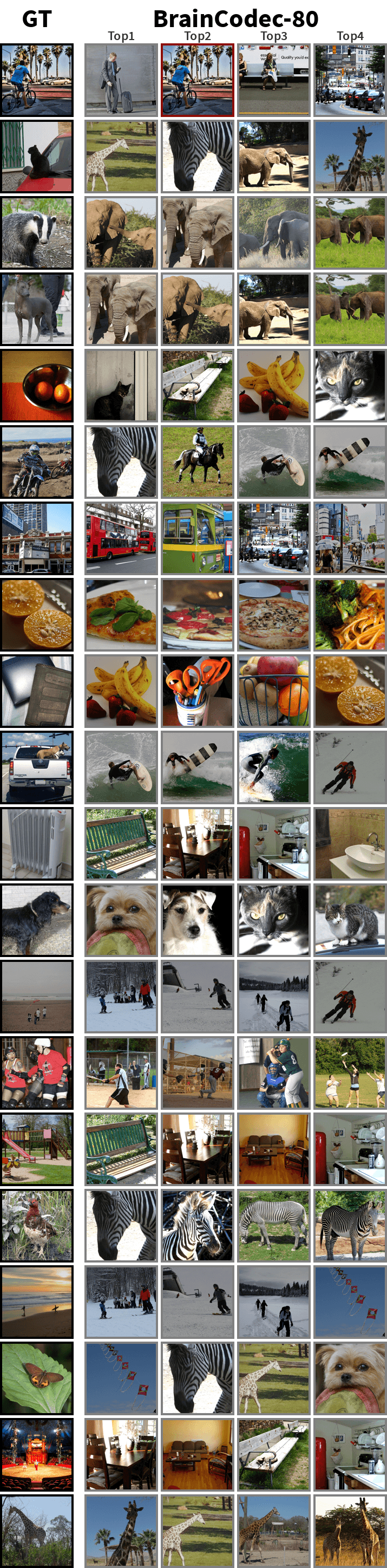}
    \caption{Subject CSI 1}
\end{subfigure}
\hfill
\begin{subfigure}{0.28\textwidth}
    \centering
    \includegraphics[width=\linewidth]{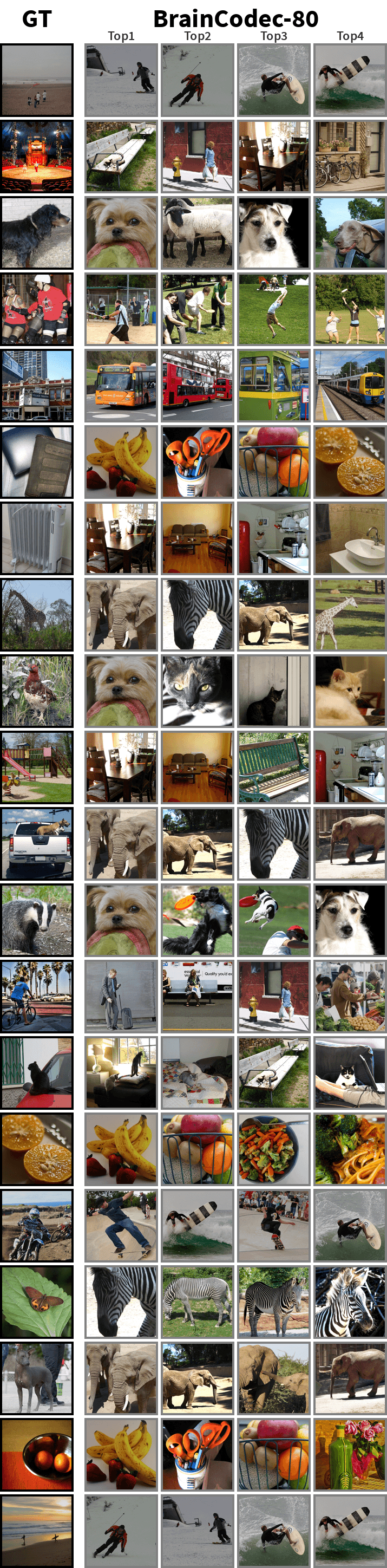}
    \caption{Subject CSI 2}
\end{subfigure}
\hfill
\begin{subfigure}{0.28\textwidth}
    \centering
    \includegraphics[width=\linewidth]{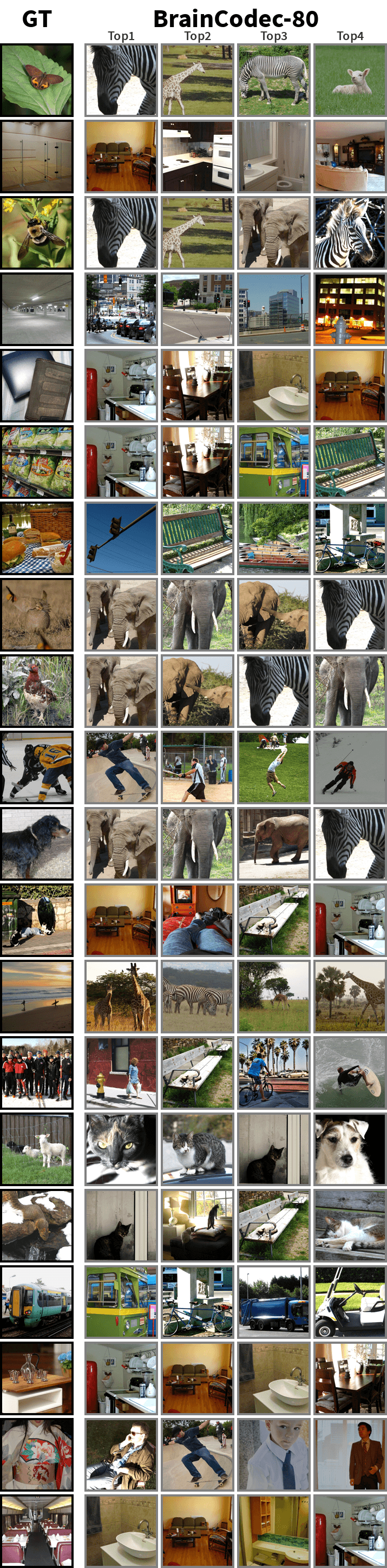}
    \caption{Subject CSI 3}
\end{subfigure}
\caption{Image retrieval results on BOLD5000 unseen subjects from fold 1 using 80 images as context. To note, since BOLD5000 provides only 20 test images, we visualize retrieval results from a pool of 500 images for rigorous evaluation.}
\end{figure}

% \pagenumbering{gobble}
% \input{suppl_table/metrics_b5k}

% \setcounter{page}{13}
\clearpage

\subsection{Ablations}
% \newpage
\label{sec:abla}
In this section, we compare different models on a variety of metrics. \textbf{PT only} indicates our model where it was only trained on synthetic data. \textbf{Inversion} is the model where we try to solve for the image embedding using gradient based optimization to recover the voxelwise activations using the stage-1 estimated voxelwise weights. For all models listed here we utilize $200$ images and brain activation patterns from the novel subject as context.
\begin{table}[h]
\centering
\caption{\textbf{Quantitative comparison on model variants and ablations.}}
\renewcommand{\arraystretch}{1.1}
\setlength{\tabcolsep}{8pt}
\begin{tabular}{lcccc}
\toprule
\textbf{Model} & \textbf{S1} & \textbf{S2} & \textbf{S5} & \textbf{S7} \\

\midrule
\multicolumn{5}{c}{\textbf{$\%$ Top-1 Accuracy ($\uparrow$)}} \\
\cmidrule(lr){1-5}
PT only                  & $3.67 \pm 0.69$& $3.24 \pm 0.71$& $2.96 \pm 0.64$& $2.68 \pm 0.63$\\
Inversion                & $1.61 \pm 0.73$& $1.39 \pm 0.86$& $2.04 \pm 0.81$& $1.90 \pm 0.77$\\
\textbf{BrainCoDec-200}  & $25.5 \pm 3.02$& $22.9 \pm 2.98$& $23.2 \pm 2.63$& $19.2 \pm 2.42$\\
BrainCoDec-200 no HO     & $28.3 \pm 3.40$& $27.1 \pm 3.21$& $29.4 \pm 3.40$& $24.0 \pm 3.36$\\
% BrainCoDec-200 no LS     & $7.14 \pm 2.06$& $4.64 \pm 2.17$& $7.60 \pm 2.33$& $5.75 \pm 2.06$\\

\midrule
\multicolumn{5}{c}{\textbf{$\%$ Top-5 Accuracy ($\uparrow$)}} \\
\cmidrule(lr){1-5}
PT only                  & $14.0 \pm 1.23$& $11.6 \pm 1.42$& $9.70 \pm 1.08$& $8.23 \pm 0.94$\\
Inversion                & $2.01 \pm 0.53$& $1.98 \pm 0.65$& $2.79 \pm 0.63$& $2.21 \pm 0.42$\\
\textbf{BrainCoDec-200}  & $56.6 \pm 3.21$& $52.4 \pm 4.08$& $55.8 \pm 2.47$& $51.2 \pm 3.50$\\
BrainCoDec-200 no HO     & $61.1 \pm2.19 $& $61.1 \pm 2.98$& $64.6 \pm 2.71$& $56.8 \pm 2.84$\\
% BrainCoDec-200 no LS     & $23.4 \pm 2.63$& $18.1 \pm 3.42$& $24.5 \pm 2.51$& $19.4 \pm 2.60$\\

\midrule
\multicolumn{5}{c}{$\%$ \textbf{Mean Rank ($\downarrow$)}} \\
\cmidrule(lr){1-5}
PT only                  & $26.63 \pm 0.93$& $27.70 \pm 0.67$& $29.63 \pm 0.87$& $30.93 \pm 1.07$\\
Inversion                & $45.87 \pm 0.87$& $46.47 \pm 0.90$& $43.97 \pm 0.77$& $46.20 \pm 1.27$\\
\textbf{BrainCoDec-200}  & $4.43 \pm 0.47$& $4.23 \pm 0.33$& $3.93 \pm 0.27$& $3.73 \pm 0.30$\\
BrainCoDec-200 no HO     & $2.67 \pm 0.27$& $3.13 \pm 0.30$& $2.50 \pm 0.13$& $3.30 \pm 0.23$\\
% \midrule
% \multicolumn{5}{c}{\textbf{Mean Rank ($\downarrow$)}} \\
% \cmidrule(lr){1-5}
% PT only                  & $79.9 \pm 2.8$& $83.1 \pm 2.0$& $88.9 \pm 2.6$& $92.8 \pm 3.2$\\
% Inversion                & $137.6 \pm 2.6$& $139.4 \pm 2.7$& $131.9 \pm 2.3$& $138.6 \pm 3.8$\\
% \textbf{BrainCoDec-200}  & $13.3 \pm 1.4$& $12.7 \pm 1.0$& $11.8 \pm 0.8$& $11.2 \pm 0.9$\\
% BrainCoDec-200 no HO     & $8.0 \pm 0.8$& $9.4 \pm 0.9$& $7.5 \pm 0.4$& $9.9 \pm 0.7$\\
% BrainCoDec-200 no LS     & $32.1 \pm 2.81$& $45.4 \pm 3.22$& $33.5 \pm 2.73$& $49.8 \pm 3.60$\\

\midrule
\multicolumn{5}{c}{\textbf{Cosine Similarity ($\uparrow$)}} \\
\cmidrule(lr){1-5}
PT only                  & $0.23 \pm 0.05$& $0.20 \pm 0.04$& $0.19 \pm 0.05$& $0.20 \pm 0.05$\\
Inversion                & $0.32\pm 0.02$& $0.30 \pm 0.02$& $0.31 \pm 0.02$& $0.31 \pm 0.07$\\
\textbf{BriancoDec-200}  & $0.81 \pm 0.01$& $0.80 \pm 0.02$& $0.79 \pm 0.03$& $0.79 \pm 0.04$\\
BrainCoDec-200 no HO     & $0.82 \pm 0.01$& $0.81 \pm 0.03$& $0.82 \pm 0.03$& $0.80 \pm 0.03$\\
% BrainCoDec-200 no LS     & $0.74 \pm 0.02$& $0.73 \pm 0.03$& $0.75 \pm 0.03$& $0.73 \pm 0.02$\\
% \textbf{BrainCodec-600}  & $0.000 \pm 0.000$ & $0.000 \pm 0.000$ & $0.000 \pm 0.000$ & $0.000 \pm 0.000$ \\
\bottomrule
\end{tabular}
\end{table}

\clearpage

% \newpage
% \input{sec/2_formatting}
% \input{sec/3_finalcopy}
{
    \small
    \bibliographystyle{ieeenat_fullname}
    \bibliography{myref}
}


\end{document}